%% file: neurips_2026.tex
\documentclass{article}

\usepackage[preprint]{neurips_2026}

\usepackage[utf8]{inputenc} 
\usepackage[T1]{fontenc}    
\usepackage{hyperref}       
\usepackage{url}            
\usepackage{booktabs}       
\usepackage{amsfonts}       
\usepackage{nicefrac}       
\usepackage{microtype}      
\usepackage{xcolor}         

\usepackage{wrapfig}
\usepackage{graphicx}
\usepackage{booktabs}
\usepackage[table,xcdraw]{xcolor}
\usepackage{colortbl} 

\newcommand{\methodname}{\textit{\textbf{Sparse Context}}}

\title{Keep The Essentials: Efficient Reference Conditioned Generation via Token Dropping}

\begin{document}

\author{
  \begin{tabular}[t]{c}
    Rishubh Parihar$^{1}$ \quad
    Ayush Raina$^{1}$ \quad 
    R. Venkatesh Babu$^{1}$ \quad 
    Or Patashnik$^{2}$ \quad
    \end{tabular}%
  \quad
  \and
  \begin{tabular}[t]{c} 
    $^1$IISc Bangalore \quad 
    $^2$Tel Aviv University \quad
  \end{tabular}%
  \vspace{2mm} \\
\textbf{Project Page:} \url{https://sparsecontext.github.io/}
}


\maketitle

\vspace{-6pt}
\begin{figure*}[h]
    \centering
    \includegraphics[width=1.0\linewidth]{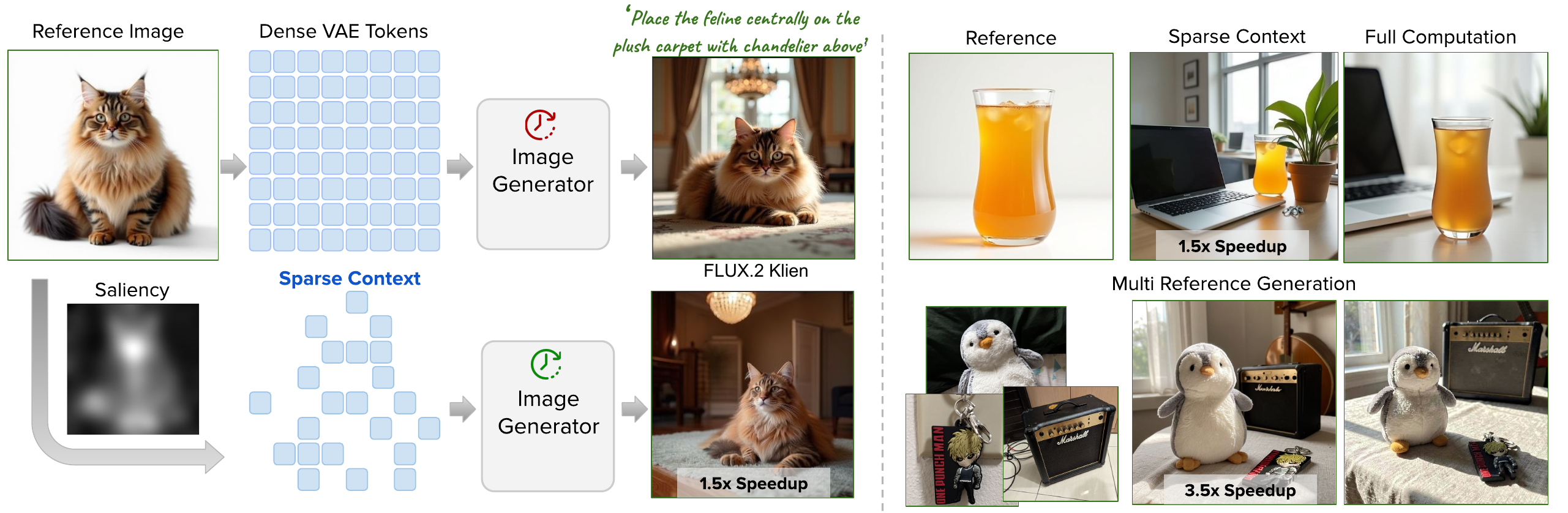}
    \vspace{-6pt}
    \caption{\methodname{} enhances the efficiency of reference-conditioned generation by replacing dense spatial grids with a smaller set of sampled tokens. Constructing sparse representations of reference images, it significantly reduces memory usage and inference time while preserving the high-quality performance of state-of-the-art generative models.}
    \label{fig:teaser}
\end{figure*}

\begin{abstract}
\input{sections/0_abstract}
\end{abstract}

\input{sections/1_introduction}

\input{sections/2_related_works}

\input{sections/3_method}
\input{sections/4_experiments}
\input{sections/5_conclusion}

\bibliographystyle{plainnat} 

\bibliography{main}


\appendix

\include{sections/supply.tex}

\newpage

\end{document}

%% file: sections/0_abstract.tex
\vspace{-6pt}
Reference-based diffusion models enable highly controllable image generation by leveraging elements from input images to guide prompt-driven synthesis. However, these models are computationally expensive in runtime, and their cost scales severely with the number of input references. While the efficiency of diffusion models has been extensively studied in the context of prompt-driven generation, it remains largely under-explored in the realm of reference-based models. This setting presents unique challenges not addressed by methods focusing solely on generation.
In particular, the wasteful representation of references as dense token grids offers significant opportunities for improvement. In this work, we present \methodname, a method for constructing sparse reference representations by retaining only a reduced subset of reference tokens. We observe that even without modifying the model, dropping a significant portion of reference tokens at inference time largely preserves its generation capabilities. To fully realize this potential, we fine-tune the model with random token dropping at varying ratios, encouraging robustness to partial reference representations. Crucially, this training strategy decouples the model from any specific token selection rule, allowing flexible control at inference time.
At inference time, instead of random dropping, we apply task-aware token selection strategies that prioritize the most informative regions of the reference images, adapting the token budget to the input and task requirements. Extensive experiments show our method achieves a $4\times$ increase in inference speed for multi-reference generation and an $2\times$ for single reference generation. Importantly, this efficiency is achieved without compromising visual quality across both spatially-aligned editing and subject-driven generation.


%% file: sections/1_introduction.tex
\section{Introduction}

\vspace{-5pt}

Diffusion-based generative models have enabled remarkable progress in text-to-image synthesis, producing high-quality and diverse outputs from natural language descriptions~\cite{Rombach_2022_CVPR, saharia2022photorealistictexttoimagediffusionmodels, dhariwal2021diffusionmodelsbeatgans}. Beyond text-only generation, many practical applications require incorporating visual references, such as editing existing images~\cite{meng2022sdeditguidedimagesynthesis, hertz2022prompt} or generating new scenes that preserve the identity or style of a given subject~\cite{ruiz2023dreambooth, gal2022imageworthwordpersonalizing}. Reference-based diffusion models address this need by conditioning the generation process on one or more input images, enabling direct visual control over the generated content~\cite{wu2025qwen, batifol2025flux}.

A common approach in recent transformer-based diffusion models (DiTs)~\cite{peebles2023scalable} is to represent reference images as tokens by encoding them with the model's VAE and concatenating the resulting latent tokens with those of the noisy input~\cite{tan2024ominicontrol, batifol2025flux}. 
This representation is inherently dense, as each reference image is mapped to a full spatial grid of tokens. Since attention operations scale quadratically with the number of tokens, this substantially increases the cost of generation. 
As reference-based generation becomes increasingly prevalent in real-world applications, where multiple high-resolution references are often used to guide synthesis, this cost can quickly become a practical bottleneck.
While efficiency has been extensively studied for prompt-driven diffusion models~\cite{lu2025toma, hu2025flashdlm, liu2025timestep}, reference-based models introduce additional computational challenges due to this dense representation, which also presents opportunities for improving efficiency.


\begin{wrapfigure}[20]{r}{0.50\textwidth}
    \vspace{-14pt}
    \includegraphics[width=1.0\linewidth]{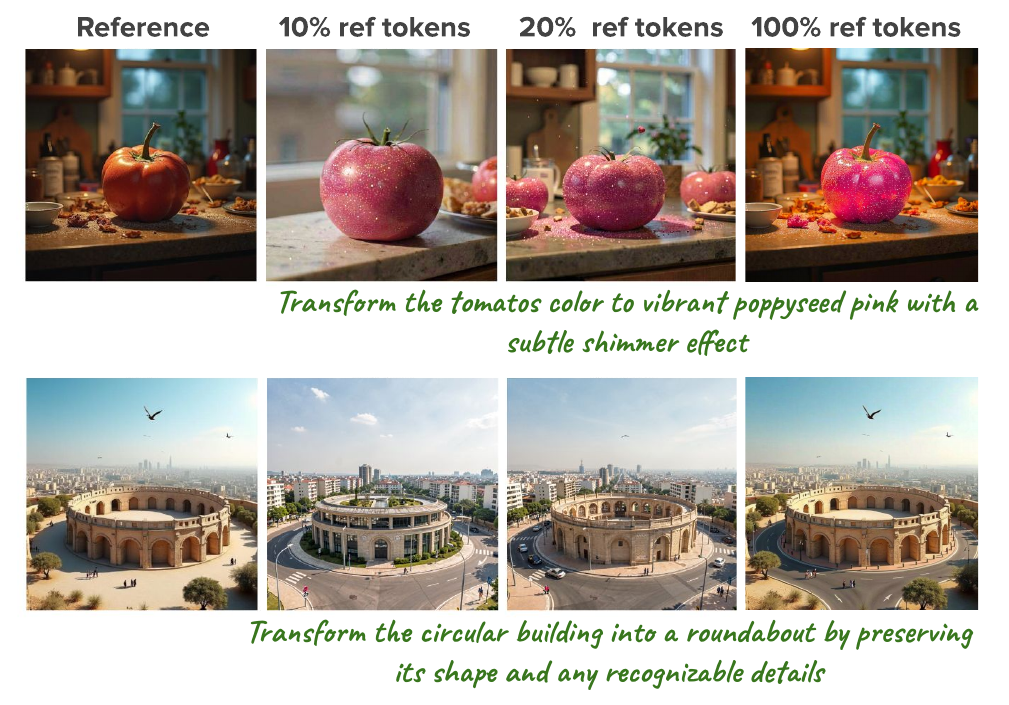}
    \vspace{-20pt}
    \caption{\textbf{Redundancy in reference tokens.} We drop a large number of reference tokens for a pretrained reference-conditioned image generator. Even when dropping $80\%$, the output resembles the coarse layout of the input scene. This finding confirms that reference tokens have high redundancy, which can be removed for efficiency.}
    \label{fig:motivation}
\end{wrapfigure}

We observe that this full-resolution representation of reference images is often unnecessary. Reference images contain regions with varying levels of detail, where many tokens carry redundant information. Surprisingly, as illustrated in Fig.~\ref{fig:motivation}, even without modifying the model, dropping a large fraction of reference tokens at inference time largely preserves generation quality with minor loss in scene details. This suggests that reference-conditioned models are inherently robust to highly sparse representations.

Motivated by this observation, we introduce \textit{\methodname}, a framework for efficient reference-based diffusion that constructs sparse reference representations. We fine-tune the model to operate on partial reference inputs by randomly dropping tokens at varying ratios during training, encouraging robustness to different subsets of reference information. Importantly, this training strategy makes the model agnostic to the specific choice of tokens, enabling flexible inference-time control. At inference, we replace random dropping with task-aware token selection strategies that prioritize the most informative reference tokens, adapting the representation to the structure of each input and the requirements of different tasks. This results in substantial reductions in computational cost while maintaining high-quality generation.

We evaluate our approach on both spatially aligned image editing and subject-driven generation tasks. By exploiting the inherent redundancy in reference-based representations, our method achieves substantial improvements in efficiency, accelerating inference by upto $2\times$ for single reference and $4\times$ for multiple reference images, while preserving fine-grained visual quality. 
Importantly, we show that our approach is orthogonal to existing efficiency techniques for diffusion models and can be readily combined with methods such as KV-caching and token merging~\cite{bolyatoken} to further improve performance.
Our results demonstrate that reference-based diffusion models can be made significantly more efficient without compromising their controllability or fidelity, highlighting the untapped efficiency potential inherent to this setting.

%% file: sections/2_related_works.tex
\vspace{-5pt}
\section{Related Works}

\vspace{-5pt}

Reference-conditioned generation extends diffusion models beyond text-only synthesis by incorporating visual inputs for improved control. 
However, adding reference images substantially increases computational cost due to the large number of conditioning tokens involved in attention. 
We review prior approaches for integrating such references into the diffusion process, along with methods for improving the efficiency of visual generative models.

\vspace{-4pt}
\subsection{Reference-based Image Generation}
\vspace{-4pt}

Standard text-to-image diffusion models~\cite{Rombach_2022_CVPR, saharia2022photorealistictexttoimagediffusionmodels} do not natively support image inputs. Consequently, dedicated methods have been developed to leverage existing images, either for image editing~\cite{meng2022sdeditguidedimagesynthesis, hertz2022prompt, Parmar_2023} or as references for subject- and style-driven generation~\cite{gal2022imageworthwordpersonalizing, ruiz2023dreambooth}.
For image editing, a prominent line of work focuses on inference-time methods that operate on the latent noise corresponding to a given image~\cite{cao2023masactrltuningfreemutualselfattention, tumanyan2022plugandplaydiffusionfeaturestextdriven, 10.1109/TPAMI.2025.3541625, https://doi.org/10.1111/cgf.15063,parihar2025zero}. When applied to real images, these approaches typically require an inversion process to map the image into the diffusion noise space~\cite{Mokady_2023_CVPR, garibi2024renoise, hubermanspiegelglas2024editfriendlyddpmnoise, kulikov2025flowedit}. To avoid this costly and often imperfect inversion process, alternative approaches generate synthetic edit pairs using such inference-time methods and subsequently train image-to-image models on the resulting data~\cite{brooks2023instructpix2pix, sheynin2024emu,parihar2026kontinuous}.
In the context of subject- and style-driven generation, early methods relied on per-subject or per-style optimization procedures~\cite{gal2022imageworthwordpersonalizing, ruiz2023dreambooth, kumari2023multi}. More advanced approaches instead learn encoders and modify the diffusion architecture to condition the denoising process on reference representations extracted from input images~\cite{guo2024pulid, xlabs-flux-ip-adapter, ye2023ipadaptertextcompatibleimage, wei2023elite,parihar2024precisecontrol}. A central challenge in training such models is the limited availability of large-scale datasets containing paired images of the same subject or style across diverse contexts.

Extended self-attention mechanisms have been shown to be effective for generating such paired data, benefiting both image editing and subject- or style-driven generation~\cite{cao2023masactrltuningfreemutualselfattention, alaluf2023crossimageattentionzeroshotappearance, avrahami2025stable, zeng2024jedi, gal2024lcmlookahead, shin2025large}. By enabling attention across multiple images, these methods promote consistency across generated outputs, for example when synthesizing the same object in different scenes.
With the rise of transformer-based diffusion models (DiTs)~\cite{peebles2023scalable}, a common strategy for incorporating image inputs is through conditioning tokens concatenated with the noisy latent tokens~\cite{tan2024ominicontrol, batifol2025flux, kumari2025generating, cai2025diffusion, agrawal2026seethrough3d}. This design allows the attention layers to enforce content preservation, whether maintaining subject identity or preserving input structure during editing. These conditioning tokens are often derived from a VAE encoder, facilitating alignment with the latent distribution used during denoising.

More recently, large-scale DiT-based foundation models have begun to natively support reference images as part of their input, integrating them through similar token concatenation mechanisms~\cite{flux-2-2025, Qwen2.5-VL, deepmind2025nanobanana}. This unifies text-to-image and image-to-image generation within a single model. Our work builds on such models to perform both image editing and subject-driven generation. We show that the attention operations, which constitute the dominant computational cost, can be made more efficient, as the relevant information can be compactly represented using a small number of conditioning tokens.





\vspace{-4pt}
\subsection{Efficient Visual Generation}
\vspace{-4pt}

Recent advances in generative models have seen a transition from U-Net-based architectures to transformer-based diffusion models (DiTs), which offer improved scalability and performance. However, this shift comes with increased computational cost, as attention operations scale quadratically with the number of tokens. In particular, in reference-conditioned settings where input images are incorporated via token concatenation, the number of tokens can grow substantially. Conditioning on a single reference image effectively doubles the token count, with additional references further amplifying this cost.

A first line of work focuses on improving the efficiency of the attention operation itself, through sparse, linearized, or hardware-aware formulations~\citep{beltagy2020longformer, choromanski2020rethinking, katharopoulos2020transformers, shen2021efficient, han2023flatten, dao2022flashattention, wang2020linformer, xia2025training}.
A second direction reduces redundant computation by caching intermediate features across layers or timesteps, with key-value (KV) caching being a prominent example, enabling reuse of previously computed representations during the denoising process~\citep{ma2024deepcache, wimbauer2024cache, ma2024learning, liu2025timestep, kahatapitiya2025adaptive, hu2025flashdlm}.
A third line of work aims to reduce the number of tokens processed by the model, either by dropping or grouping tokens. Initially explored in Vision Transformers for image understanding~\citep{bolyatoken, kim2024token}, these approaches have more recently been adapted to generative settings with DiT architectures~\citep{bolya2023token, lu2025toma, wu2025importance, hu2024token}.

While these methods are designed for general generative settings, reference-conditioned generation offers additional opportunities for efficiency. In this setting, the source image is fully observed, and its information can be compactly encoded rather than processed at full token resolution. In our work, we leverage this property to compress the reference into a small set of spatially-aware conditioning tokens, significantly reducing the cost of attention while preserving the relevant information.






%% file: sections/3_method.tex
\section{Method}

\subsection{Preliminaries} 

\textbf{Diffusion Models.} Diffusion or flow based models~\cite{song2022denoisingdiffusionimplicitmodels, lipmanflow} learn to map from Gaussian noise distribution to the image distribution. In the forward diffusion process, the input image $x~\in\mathcal{X}$ is corrupted with a Gaussian noise $\epsilon \sim \mathcal{N}(0,I)$ to obtained noisy version of the sample $x_t = x + \alpha_t \epsilon$, where and $\alpha_t$ follows a predefined noise schedule. To learn the reverse mapping, a time-conditioned model $\epsilon_\theta(x_t,t)$ is trained to predict the added noise. For conditional generation the denoising network is conditioned with additional input signal $y$ i.e. $\epsilon_\theta(x_t,t,y)$, such as text prompt for text-to-image generation or images for reference conditioned generation.



%

\textbf{Reference conditioned DiTs.} State-of-the-art diffusion models~\cite{flux} are implemented with Diffusion Image Transformer (DiT)~\cite{peebles2023scalable}, where the images $x\in\mathcal{X}$ are encoded in VAE latent space and tokenized to obtain patch-level representations $z\in\mathcal{Z}$. These visual tokens are noised and processed by a sequence of DiT blocks, where self-attention enables interactions across all tokens. This token-based formulation naturally supports conditioning on image references, where the reference image is first tokenized $y$ and its tokens are concatenated to the noisy image tokens $z_t$. The self-attention mechanism enables strong conditioning between the reference images and the noisy latents enbaling high-quality reference conditional generation. 

\begin{figure}[t]
    \centering
    \includegraphics[width=\linewidth]{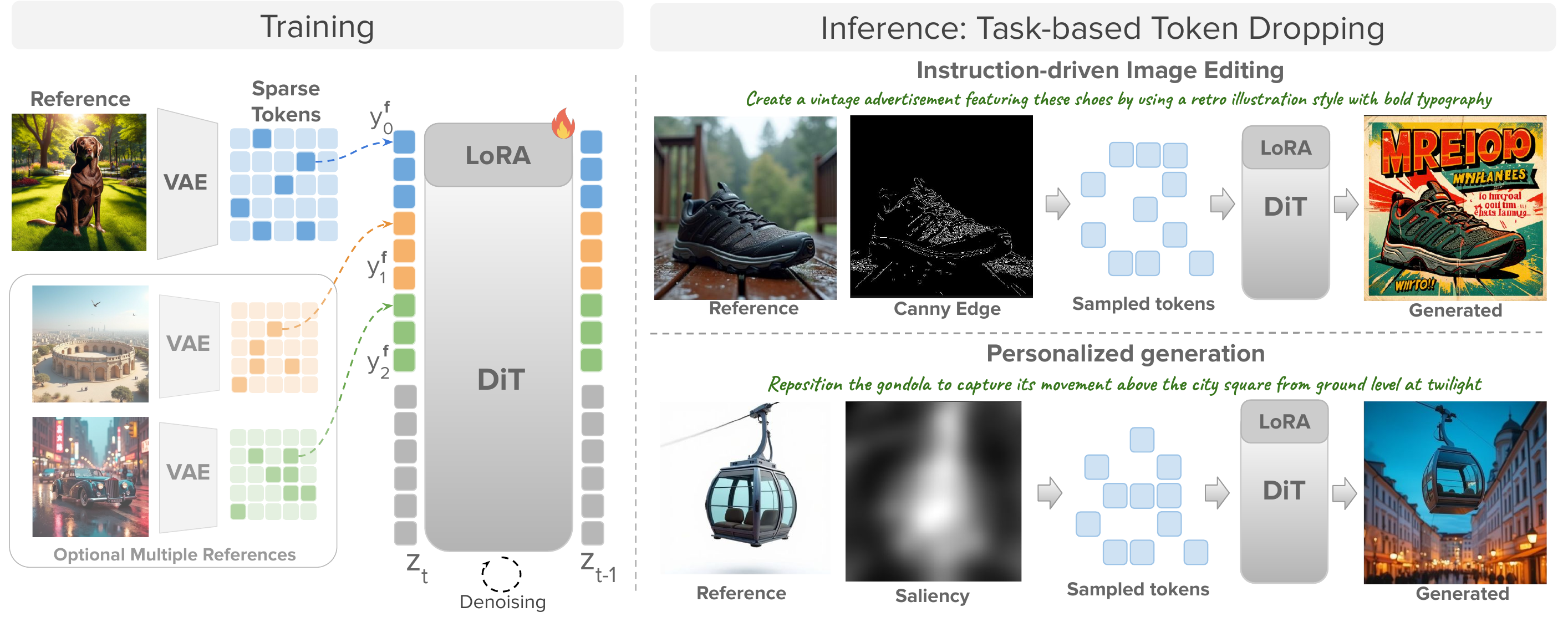}
    \caption{\methodname\; \textbf{Overview}. \textbf{Training:} During training, we randomly drop the tokens from reference images with a keep fraction of $f\in(0.05,0.25)$ to obtain their sparse token representation $y^f_i$ for conditioning to denoise the image tokens $z_t$. \textbf{Inference:} During inference, user can randomly drop the reference tokens based on their budget and condition the DiT on the sparse tokens for reference based generation. Further, different task-specific token dropping strategies can be used for highest quality given a fixed compute budget. For instance sampling more tokens from the edge regions to preserve image structure during image editing or sampling from salient regions for personalization.}
    \label{fig:method_oveview}
\end{figure}

\subsection{Redundancy in Reference Image Tokens}
A reference conditioned generator takes as input a text prompt and a one or multiple $n$ reference images $\{y_i\}_{i=1}^n$, where $y_i\in\mathcal{X}$ to produces an edited image. The reference tokens are appended to the image latents $x_t$ and processed by the DiT model. While powerful, this conditioning mechanism significantly increases computational cost, as self-attention scales quadratically with the number of input tokens. This issue becomes particularly severe when multiple references are used as the number of reference tokens increase linearly with number of reference images.  

This raises a natural question: \textit{do we actually need all reference image tokens?} To investigate this, we perform a simple experiment using the state-of-the-art generator FLUX.2-Klien~\cite{flux-2-2025} in which we randomly drop tokens from a reference image $y_i$ with a keep fraction of $f\in\{0.10,0.20\}$, resulting in a sparse reference representation $y_i^f$ at inference time. As shown in Fig.~\ref{fig:motivation}, even after removing $0.8$ fraction of the reference tokens, the generated image still preserves the overall layout of the reference, although fine object and scene details are largely lost. This is a training-free setting, which we refer to as \textit{Na\"{i}ve Context Pruning}. This observation suggests that a significant portion of the reference tokens is redundant and can be removed. Bridging the remaining gap in fine details could therefore enable substantial acceleration at inference.

\subsection{\methodname}
We hypothesize that the performance gap between full-reference and partial-reference generation primarily arises from a train–test distribution mismatch induced by sparse reference conditioning at inference time. While the model is trained to condition on the full set of tokens $y_i$ extracted from the reference image, it is used at inference time with only a subset of them $y_i^f$.
To mitigate this mismatch, we fine-tune the generator to operate under sparse reference conditioning. During training, we randomly drop tokens with a keep fraction of $f\in(0.05,0.25)$ of the reference tokens before concatenating them with the complete noisy target tokens $z_t$ as shown in Fig.~\ref{fig:method_oveview}-left. The self-attention mechanism enables the model to effectively condition on the remaining reference tokens despite the reduced input. 

\begin{figure}[]
    \centering
    \includegraphics[width=\linewidth]{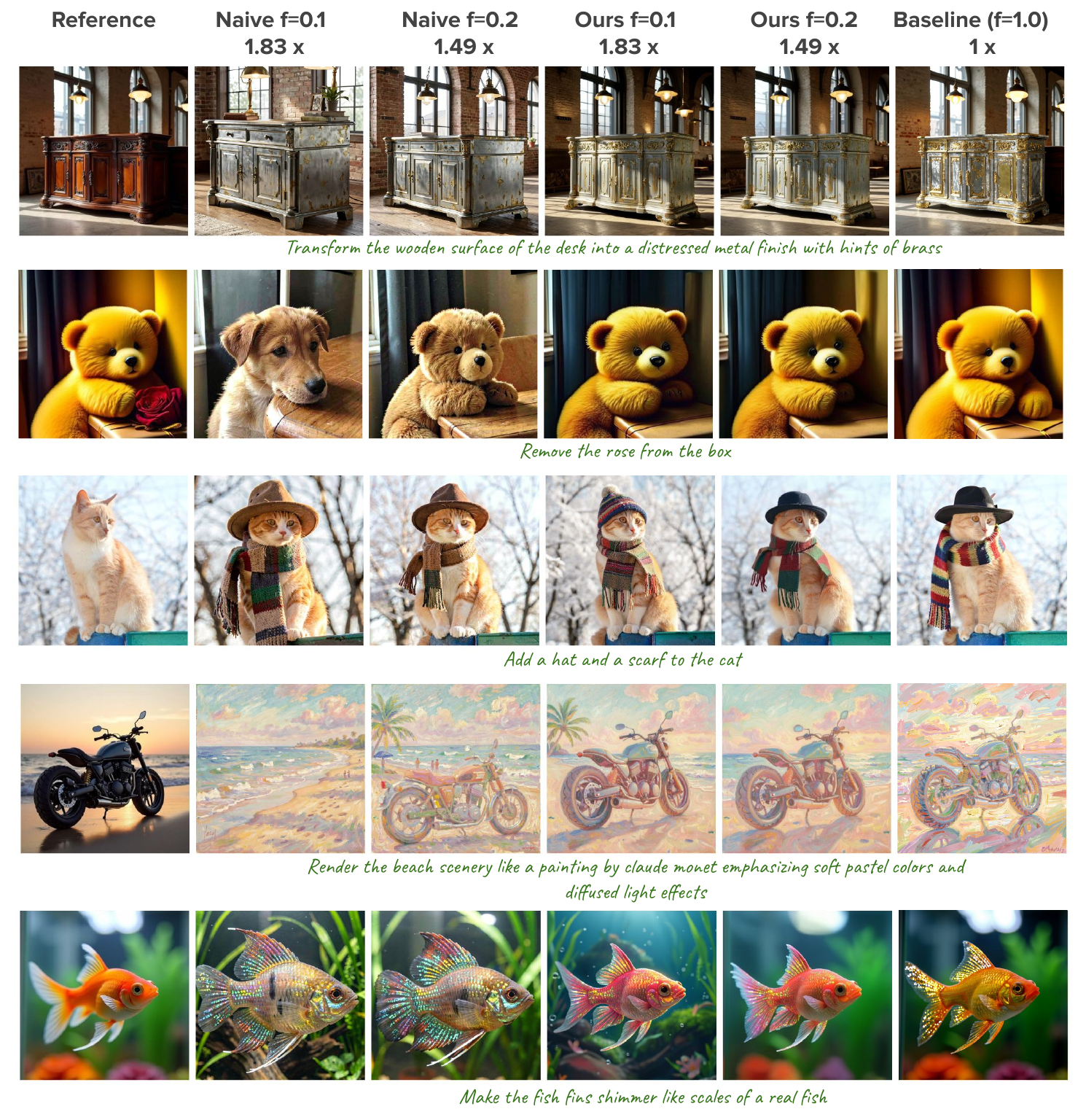}
    \vspace{-6mm}
    \caption{\textbf{Image Editing Results.} We perform instruction based image editing with diverse editing prompts. \methodname{} preserves the structure of the reference image during editing while accurately performing the described edit.}
    \vspace{-6mm}
    \label{fig:qual-editing}
\end{figure}

In practice, we fine-tune a state-of-the-art reference-conditioned generator~\citep{flux-2-2025} using a lightweight LoRA on a curated dataset for reference-based image generation (Sec.~\ref{subsec:dataset}). This adaptation substantially improves scene structure and identity preservation compared to the \textit{Na\"{i}ve Context Pruning} baseline (Fig.~\ref{fig:qual-editing}), supporting our hypothesis.
Using a randomly sampled drop ratio during training encourages the model to operate robustly across different levels of sparsity in the reference tokens and enables flexible control over the inference-time compute budget while preserving essential reference information. At inference time, reference tokens can be randomly dropped according to the desired compute budget, allowing a direct trade-off between efficiency and generation quality.


\subsection{Task-aware Token Dropping}

Reference-conditioned models are used in a variety of settings, such as image editing, which requires spatially aligned generation, and personalization, where subjects from reference images are either placed in new environments or combined into novel compositions. Our training strategy of randomly dropping tokens is agnostic to task-specific inductive biases, enabling flexible and general use. In particular, it facilitates the design of task-aware token dropping strategies at inference time across a wide range of reference-conditioned generation tasks. We explore two representative tasks below and propose task-specific token dropping strategies for each.

\paragraph{Instruction-driven Image Editing.}
Given a reference image and edit instructions in the form of a text prompt, the task is to generate an edited image that follows the specified instructions. A key requirement for consistent image editing is preserving the structure of the reference image while precisely modifying the desired aspects of the scene. In this regard, edge maps provide informative cues about scene structure, as they highlight object boundaries. To better preserve the input structure, we propose an edge-aware token sampling strategy, where more tokens are sampled from edge regions compared to uniform regions. 
Specifically, we first extract a Canny edge map from the reference

\begin{wrapfigure}[]{r}{0.55\textwidth}
    \vspace{-2mm}
    \includegraphics[width=1.0\linewidth]{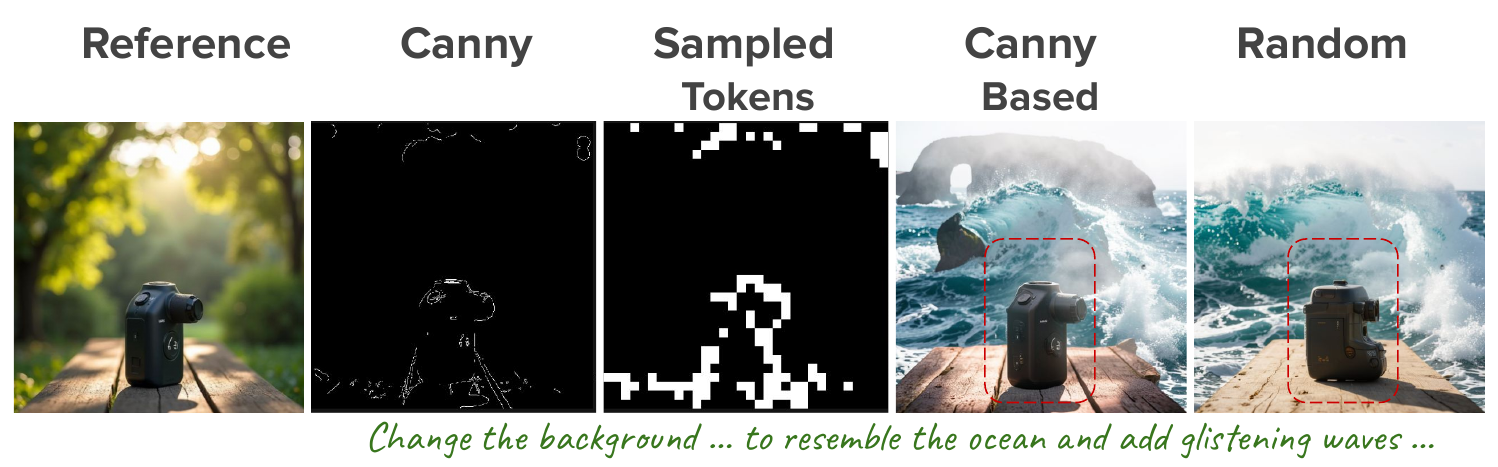}
    \vspace{-6mm}
    \caption{\textbf{Token selection for image editing.} During inference, we use Canny edge map to concentrate token selection on structural boundaries rather than random sampling. This edge-prioritized strategy more effectively preserves the underlying image structure and scene identity during the editing process.}
    \label{fig:canny-ablate}
    \vspace{-8mm}
\end{wrapfigure} 

image and use it as a probability distribution to sample reference tokens. As shown in Fig.~\ref{fig:canny-ablate}, this approach better preserves scene structure than random dropping while maintaining accurate edits. Under the same token budget, edge-based sampling significantly improves structural consistency.

\begin{wrapfigure}[]{r}{0.55\textwidth}
    \vspace{-8mm}
    \includegraphics[width=1.0\linewidth]{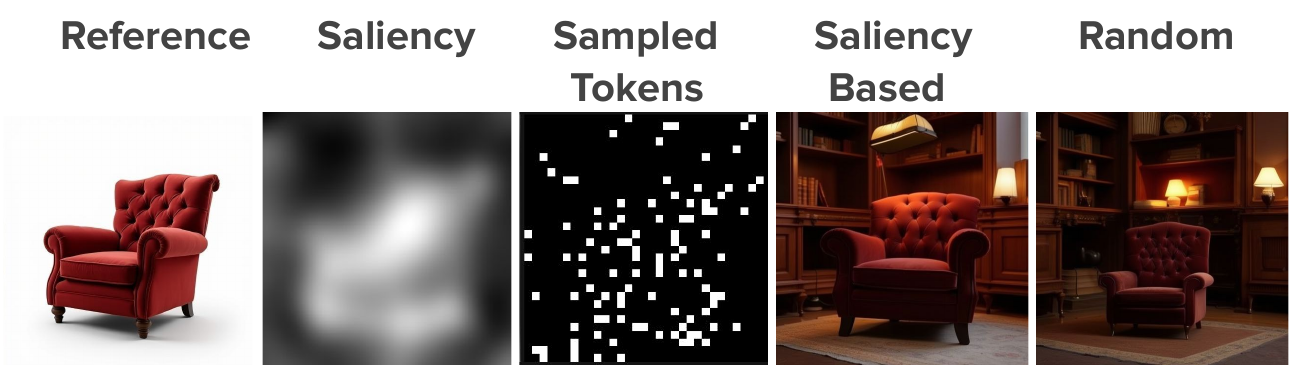}
    \vspace{-6mm}
    \caption{\textbf{Token Selection for Personalization.} We utilize a saliency map to localize the dominant scene object, allowing the selection process to concentrate on informative regions. By prioritizing these salient tokens, our method more effectively extracts fine-grained object details, leading to significantly higher identity preservation critical for high-fidelity image personalization.}
    \vspace{-4mm}
    \label{fig:saliency-ablate}
\end{wrapfigure} 

\paragraph{Personalization.}
Given one or a few reference images, the task of personalization is to generate subjects from the reference images in new compositions. A key requirement for accurate personalization is preserving the identity of the subjects from the source images. In typical real-world images, the subject occupies only a small portion of the pixels. To this end, we use image saliency as an approximation of the subject mask in the reference image and employ it to perform weighted sampling of source tokens. As shown in Fig.~\ref{fig:saliency-ablate}, selecting tokens based on saliency significantly improves subject identity in the generated scene. Alternatively, segmentation maps of the input image can be used for more accurate subject localization.

\subsection{Dataset}
\label{subsec:dataset}
To train our model with token dropping, we construct a diverse dataset covering the main reference-conditioned tasks: instruction-driven editing and personalization. For instruction-driven editing, we use a $30K$ subset of HQ-Edit~\citep{huihq}, which contains high-quality source images and edit instructions. We regenerate the edited images using Flux-Klien2~\citep{flux-2-2025}, as the original edited images in HQ-Edit are not sufficiently high fidelity.
For personalization, we use a $20K$ subset of the Subjects-200K dataset~\cite{tan2024ominicontrol}, which contains pairs of reference images of subjects on plain backgrounds and their corresponding compositions. In addition, for multi-reference generation, we generate $13K$ images, each containing compositions of $2$-$6$ objects, using the CustomDiffusion-105 dataset~\cite{kumari2023multi}. Additional details about the dataset are provided in the appendix.

%% file: sections/4_experiments.tex
\section{Experiments}


\paragraph{Implementation Details.} 
We perform all our experiments using the FLUX-2-Klein-9B~\cite{flux-2-2025} base model. All images used for training and evaluation are at a resolution of $512 \times 512$. Training is performed via efficient LoRA fine-tuning with rank $16$, for a total of $15K$ iterations. We employ a curriculum training strategy, where the first $10K$ iterations use samples with a single reference image, and the remaining $5K$ iterations use a mixture of single-reference and multi-reference samples from our custom dataset, as described above. We train with an effective batch size of $8$ on two NVIDIA A100 GPUs.

\begin{figure}[t]
    \centering
    \includegraphics[width=1.0\linewidth]{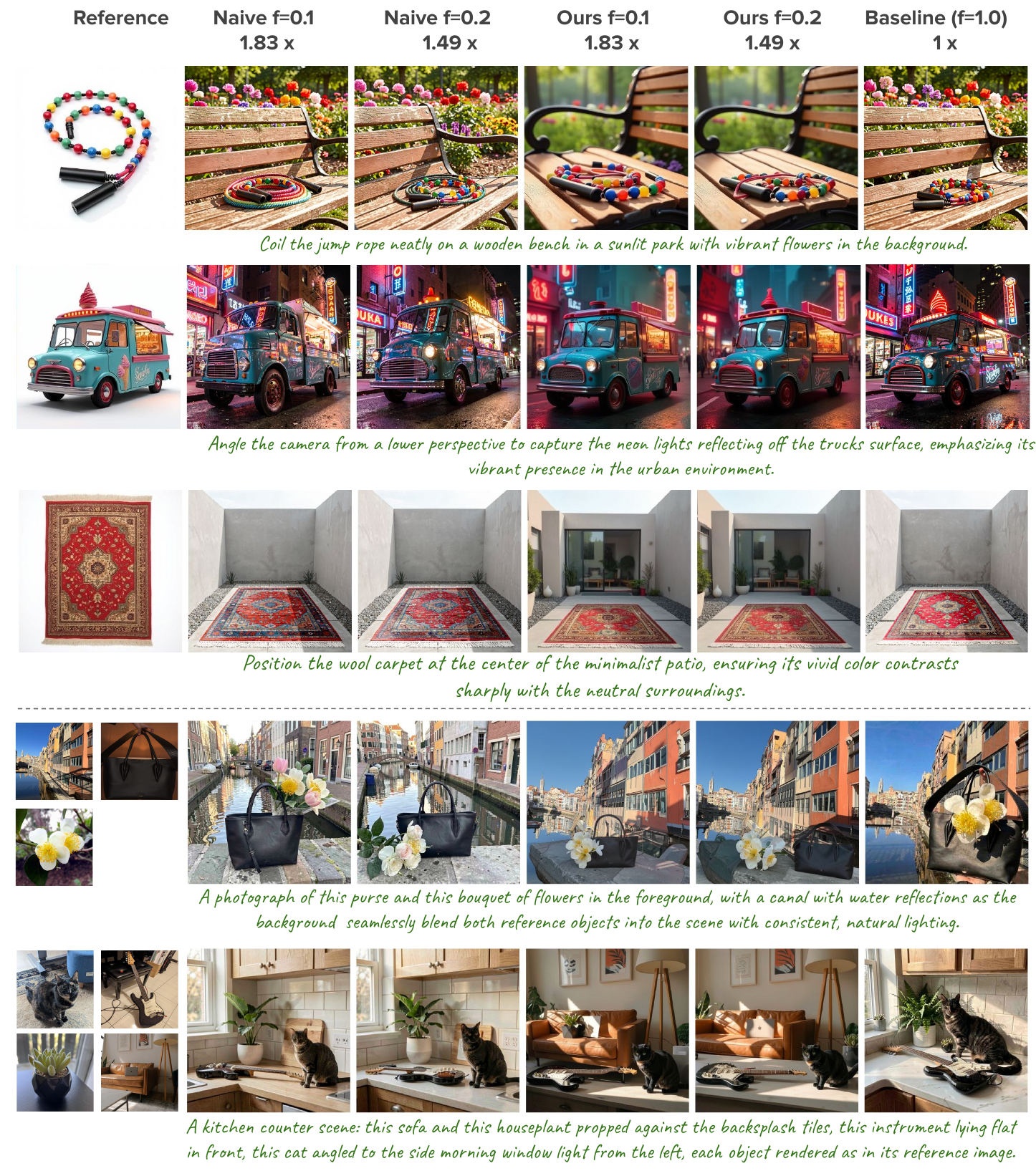}
    \vspace{-6mm}
    \caption{\textbf{Image Personalization Results.} Our method preserves the object identity well even with only keeping $10\%$ of reference tokens ($f=0.1$). Na\"{i}ve token dropping fails in preserving fine-grained object details such as colors of the rope or carpet pattern.}
    \vspace{-4mm}
    \label{fig:qual-personalization}
\end{figure} 

\begin{table}[b]
\caption{Evaluation of generation quality with different reference token fractions $f$.}
\resizebox{\textwidth}{!}{%
\begin{tabular}{@{}lccccccc|cccccc@{}}
\toprule
 & \multicolumn{7}{c|}{\textbf{Instruction Driven Image Editing}} & \multicolumn{6}{c}{\textbf{Reference Based Personalization}} \\
\midrule
 & \textbf{Tok.} $f$ & \textbf{LPIPS} $\downarrow$ & \textbf{CLIP-I} $\uparrow$ & \textbf{DINO} $\uparrow$ & \begin{tabular}[c]{@{}c@{}}\textbf{KID} $\downarrow$ \\ \small{($\times10^{-3}$)}\end{tabular} & \textbf{CLIP T-I} $\uparrow$ & \textbf{Speedup} $\uparrow$ &
\textbf{Tok.} $f$ & \textbf{CLIP-I} $\uparrow$ & \textbf{DINO} $\uparrow$ & \textbf{CLIP T-I} $\uparrow$ & \begin{tabular}[c]{@{}c@{}}\textbf{KID} $\downarrow$ \\ \small{($\times10^{-3}$)}\end{tabular} & \textbf{Speedup} $\uparrow$ \\
\midrule
\rowcolor[HTML]{D9D9D9}
Baseline & 1.00 & 0.421 & 0.826 & 0.761 & 1.36 & 0.319 & 1.00x & 1.00 & 0.708 & 0.412 & 0.287 & 28.4 & 1.00x \\ \midrule
Na\"{i}ve & 0.05 & 0.740 & 0.675 & 0.446 & \textbf{4.4} & 0.297 & 2.06x & 0.05 & 0.654 & 0.318 & \textbf{0.286} & 28.6 & 2.06x \\
Ours      & 0.05 & \textbf{0.665} & \textbf{0.772} & \textbf{0.660} & 5.1 & \textbf{0.304} & 2.06x & 0.05 & \textbf{0.735} & \textbf{0.435} & 0.275 & \textbf{23.6} & 2.06x \\ \midrule
Na\"{i}ve & 0.10 & 0.719 & 0.723 & 0.553 & 3.4 & 0.309 & 1.83x & 0.10 & 0.681 & 0.360 & \textbf{0.287} & 27.8 & 1.83x \\
Ours      & 0.10 & \textbf{0.628} & \textbf{0.796} & \textbf{0.703} & \textbf{3.3} & \textbf{0.311} & 1.83x & 0.10 & \textbf{0.748} & \textbf{0.452} & 0.274 & \textbf{22.7} & 1.83x \\ \midrule
Na\"{i}ve & 0.20 & 0.687 & 0.762 & 0.631 & 2.9 & \textbf{0.316} & 1.49x & 0.20 & 0.692 & 0.390 & \textbf{0.288} & 27.0 & 1.49x \\
Ours      & 0.20 & \textbf{0.576} & \textbf{0.817} & \textbf{0.738} & \textbf{2.3} & 0.314 & 1.49x & 0.20 & \textbf{0.754} & \textbf{0.455} & 0.274 & \textbf{23.0} & 1.49x \\ \bottomrule
\end{tabular}%
}
\label{tab:quantiative}
\end{table} 

\paragraph{Evaluation Dataset.}
For image editing, we use the standard and diverse PIE-Bench~\cite{ju2023pnp} benchmark, which consists of $700$ source images, edit instructions, and edited images covering a wide range of editing operations, including stylization, object insertion/removal, and object replacement. For personalization, we evaluate on a subset of $200$ examples from Subject-200K for single-reference personalization and on $500$ compositional scenes from CustomDiffusion-105 using multiple reference images ($2$--$6$ references per scene).

\paragraph{Evaluation Metrics.}
We report LPIPS~\cite{zhang2018unreasonable}, CLIP-Image (CLIP-I)~\cite{radford2021learning}, and DINO~\cite{oquab2023dinov2} similarity scores between the reference and generated images to measure structure and identity preservation. For personalization, we do not report LPIPS, as the generated outputs are not expected to be pixel-aligned with the reference images. To evaluate text-image alignment, we report the CLIP text-image similarity score (CLIP T-I). For PIE-Bench, we use the target image captions to compute the CLIP alignment score, while for personalization, we use the generation prompts. Finally, we report relative inference speedups averaged over $100$ runs on a single NVIDIA $A100$ GPU, compared to the base model without token dropping.


\subsection{Instruction-driven Image Editing}
\vspace{-2mm}
We present qualitative results in Fig.~\ref{fig:qual-editing}, where we compare our method with the \textit{Na\"{i}ve Context Pruning} baseline (\textbf{Na\"{i}ve}) and the full-reference generation using all reference tokens (\textbf{Baseline}). Our method preserves the scene structure even when using only $10\%$ of the reference tokens, while effectively performing the intended edits. In contrast, the \textit{Na\"{i}ve Context Pruning} baseline corrupts the scene identity, as seen in the wardrobe in the first row and the teddy bear appearance in the second row. On the efficiency front, we achieve a $1.49\times$ speedup using only $20\%$ of the reference tokens, while maintaining quality comparable to that of generation with the full set of reference tokens. We also observe marginal improvements in scene structure when increasing the token budget from $10\%$ to $20\%$. This provides users with inference-time control over the compute budget while still achieving consistent image edits.
We present quantitative comparisons in Tab.~\ref{tab:quantiative}, where our method demonstrates significantly better identity preservation, reflected by lower LPIPS and higher DINO and CLIP-I scores compared to \textit{Na\"{i}ve Context Pruning}, while maintaining similar text-image alignment (CLIP T-I). Furthermore, our method achieves more than a $2\times$ speedup at a $0.05$ keep fraction compared to the full-reference baseline using all reference tokens.

\subsection{Personalization} 
\paragraph{Single reference.} 
We present qualitative personalization results in Fig.~\ref{fig:qual-personalization}. Our method achieves highly consistent personalization results, where the reference object is naturally integrated into the generated scene according to the text prompt, while preserving fine-grained subject details, such as the painting in the first row and the hood of the car in the second row, despite using only a small fraction of the reference tokens. Our saliency-based token sampling successfully concentrates the representation on the object regions. 
In contrast, applying the same sampling strategy to the \textit{Na\"{i}ve Context Pruning} baseline significantly distorts the subject identity, for example by incorrectly introducing human figures into paintings or altering vehicle structures. Quantitative results in Tab.~\ref{tab:quantiative} show that our method maintains strong identity preservation, reflected by high CLIP-I and DINO scores, while preserving text-image alignment. In practice, a CLIP T-I score above $0.26$ generally indicates good prompt adherence, which is also evident in the qualitative results in Fig.~\ref{fig:qual-personalization}.

\begin{wrapfigure}[]{r}{0.55\textwidth}
    \vspace{-2mm}
    \includegraphics[width=1.0\linewidth]{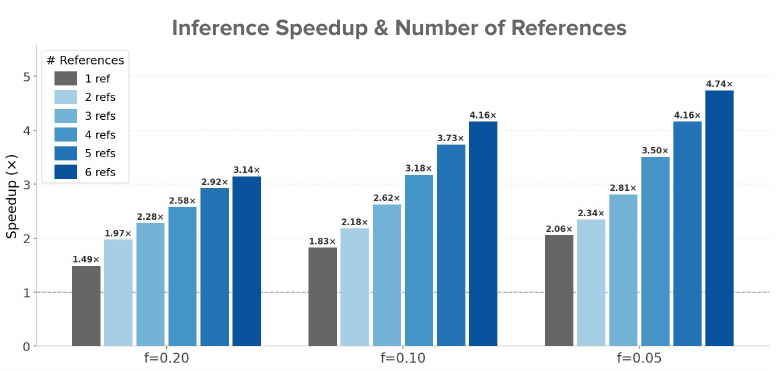}
    \vspace{-6mm}
    \caption{Speedup for multi-reference personalization across different drop fraction.}
    \vspace{-2mm}
    \label{fig:speedups-multi-reference}
\end{wrapfigure} 

\paragraph{Multiple references.} As shown in Fig.~\ref{fig:qual-personalization}, given multiple reference images, our method generates natural scene compositions while accurately preserving object identities, such as the yellow flowers and background in the fourth row and the sofa in the last row. Since quantitatively measuring identity preservation in multi-reference generation is challenging, we evaluate the generated images using text-image alignment metrics (reported in the appendix) and present inference speedups across varying numbers of reference images in Fig.~\ref{fig:speedups-multi-reference}. Our inference-time speedups scale with the number of reference images, as reference tokens occupy the majority of the self-attention computation. For example, with $n=6$ reference images, $86\%$ of the tokens correspond to reference images, while only $14\%$ correspond to the target image. \methodname{} effectively removes redundant reference tokens, leading to substantial speedups. In particular, for more than five reference images, we achieve over $4\times$ speedups with keep fractions of $f=0.10$ and $f=0.05$.

\subsection{Integration with other efficiency methods}

Our proposed method for efficient generation is simple and can easily be plugged into existing efficiency approaches. Here we show the integration of our method with two commonly used inference-time efficiency approaches: Token Merging and KV Caching.

\textbf{Token Merging:} We also integrate our method with a modification of the Token Merging~\cite{bolya2023token} method. Given a set of reference tokens, Token Merging aims to merge similar tokens and hence reduce the total number of tokens. In our context of reference-based generation, we perform token merging only once after encoding the reference images through the VAE encoder in the spirit of our one-shot token reduction method. We present the results of performing token merging (ToMe) on the pretrained model and token merging with our fine-tuned model in Tab.~\ref{tab:token-merging}.  When combined with our pretrained model, the image preservation and text-to-image alignment is improved for both the image editing and personalization tasks. The key observation is that even though our fine-tuning is done with random token dropping, it effectively improves the inference time token reduction through merging, indicating strong compatibility with existing inference time token reduction methods.

\begin{table}
\caption{Integration of SparseContext with inference time Token merging~\cite{bolya2023token}.}
\resizebox{\textwidth}{!}{%
\begin{tabular}{@{}lccccccc|cccccc@{}}
\toprule
 & \multicolumn{7}{c|}{\textbf{Instruction Driven Image Editing}} & \multicolumn{6}{c}{\textbf{Reference Based Personalization}} \\
\midrule
 & \textbf{Tok.} $f$ & \textbf{LPIPS} $\downarrow$ & \textbf{CLIP-I} $\uparrow$ & \textbf{DINO} $\uparrow$ & \begin{tabular}[c]{@{}c@{}}\textbf{KID} $\downarrow$ \\ \small{($\times10^{-3}$)}\end{tabular} & \textbf{CLIP T-I} $\uparrow$ & \textbf{Speedup} $\uparrow$ &
\textbf{Tok.} $f$ & \textbf{CLIP-I} $\uparrow$ & \textbf{DINO} $\uparrow$ & \textbf{CLIP T-I} $\uparrow$ & \begin{tabular}[c]{@{}c@{}}\textbf{KID} $\downarrow$ \\ \small{($\times10^{-3}$)}\end{tabular} & \textbf{Speedup} $\uparrow$ \\
\midrule
\rowcolor[HTML]{D9D9D9}
Baseline & 1.00 & 0.421 & 0.826 & 0.761 & 1.36 & 0.319 & 1.00x & 1.00 & 0.708 & 0.412 & 0.287 & 28.4 & 1.00x \\ \midrule
ToMe & 0.05 & 0.730 & 0.651 & 0.412 & 4.86 & 0.290 & 2.10x & 0.05 & 0.641 & 0.301 & \textbf{0.285} & 30.2 & 2.10x \\
Ours + ToMe & 0.05 & \textbf{0.614} & \textbf{0.769} & \textbf{0.675} & \textbf{2.51} & \textbf{0.306} & 2.10x & 0.05 & \textbf{0.721} & \textbf{0.411} & 0.273 & \textbf{26.7}  & 2.10x \\ \midrule
ToMe & 0.10 & 0.690 & 0.707 & 0.538 & 2.64 & 0.304 & 1.76x & 0.10 & 0.664 & 0.341 & \textbf{0.287} & 27.4 & 1.76x \\
Ours + ToMe & 0.10 & \textbf{0.570} & \textbf{0.798} & \textbf{0.733} & \textbf{1.42} & \textbf{0.314} & 1.76x & 0.10 & \textbf{0.739} & \textbf{0.436} & 0.273 & \textbf{25.4}  & 1.76x \\ \midrule
ToMe & 0.20 & 0.637 & 0.767 & 0.653 & 1.48 & 0.315 & 1.40x & 0.20 & 0.687 & 0.377 & \textbf{0.288} & 27.5 & 1.40x \\
Ours + ToMe & 0.20 & \textbf{0.529} & \textbf{0.818} & \textbf{0.767} & \textbf{1.36} & \textbf{0.316} & 1.40x & 0.20 & \textbf{0.747} & \textbf{0.448} & 0.274 & \textbf{25.1} & 1.40x \\ \bottomrule
\end{tabular}%
}
\label{tab:token-merging} 
\end{table}

\textbf{KV-cache:}
KV-cache~\cite{radfordlanguage} is one of the widely used techniques to improve the efficiency of attention-based models. The idea is to store the key and values from the first computation and later

\begin{wrapfigure}[]{r}{0.55\textwidth}
    \vspace{-5mm}
    \includegraphics[width=1.0\linewidth]{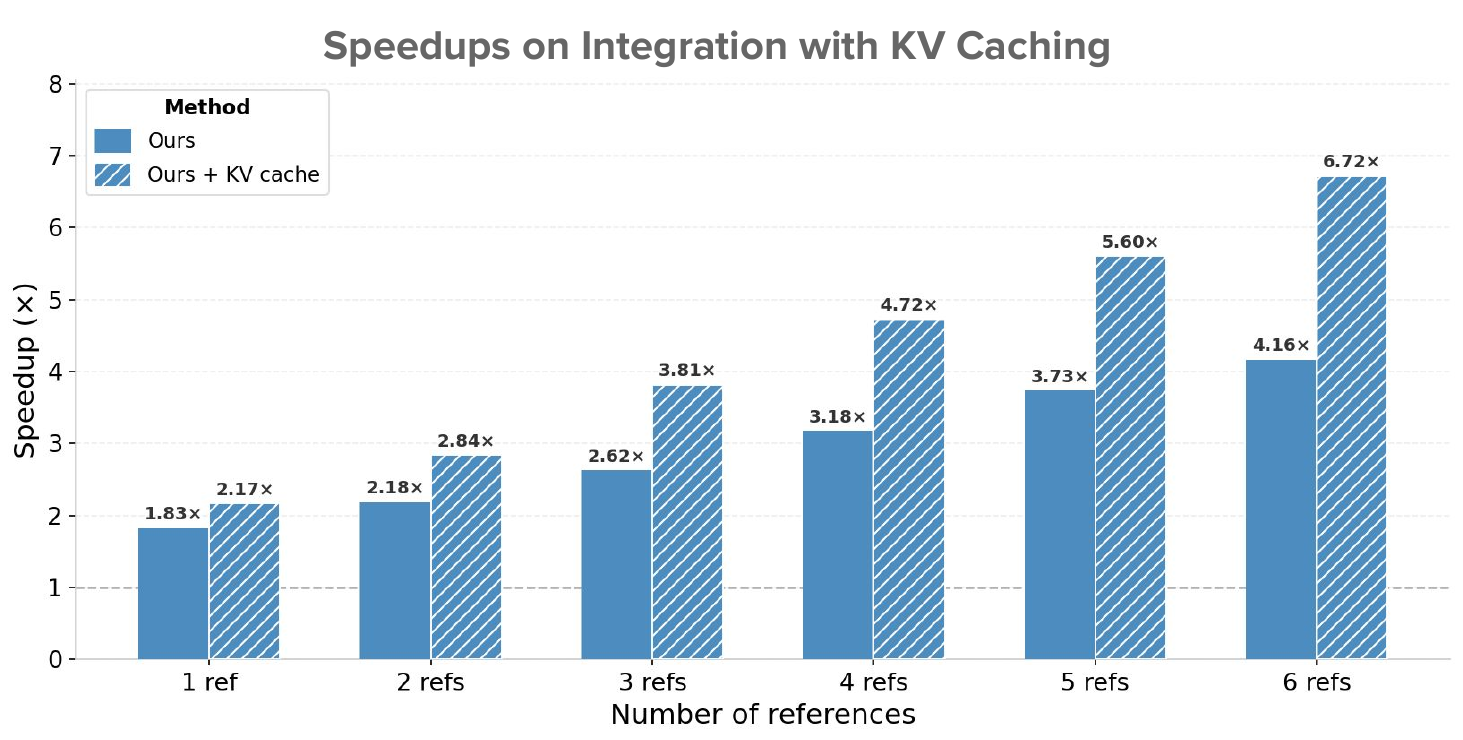}
    \vspace{-6mm}
    \caption{Our method is complementary to KV-caching and can be easily integrated to achieve further speedup gains for reference-conditioned generation.}
    \label{fig:kv-cache-ablation}
\end{wrapfigure} 

retrieve the stored key and values for the attention computation. In the context of reference-based image generation with diffusion models, the KV of the reference tokens is stored during the first denoising step and then retrieved and used for subsequent steps. This provides substantial inference time speedups, and similar to our approach, the gains improve with the number of reference images. Our method and KV-cache explore complementary directions for efficiency, and KV-cache can be integrated with our method to achieve compounded efficiency gains. To this end, we implement the KV-cache on our selected reference tokens for the multi-reference generation task with $f=0.10$, and the results are presented in Fig.~\ref{fig:kv-cache-ablation}. Notably, our method, when integrated with KV-cache, achieves substantial speedup up to $6.72\times$ when prompted with $6$ reference images.  

%% file: sections/5_conclusion.tex
\section{Conclusion and Discussion}
We propose \methodname{}, a simple token-dropping strategy for efficient reference-conditioned image generation. We show that most tokens originating from reference images are redundant and can be discarded, significantly reducing the quadratic cost of the self-attention mechanism. However, na\"ively dropping tokens at inference time distorts scene identity due to the mismatch between the training and inference distributions. To address this, we perform lightweight post-training on a curated dataset consisting of two reference-conditioned generation tasks.
The resulting model is substantially more efficient at inference time, achieving up to $5\times$ speedups for multi-reference generation. Furthermore, our generalized retraining strategy enables flexible inference-time token dropping schemes tailored to specific tasks, such as saliency-guided token selection for reference-based generation. Our method is complementary to existing efficiency techniques, such as KV-caching, and can be combined with them for additional gains. More broadly, our findings highlight a significant source of redundancy in contemporary DiT-based image and video generation models, which are otherwise extremely expensive at high resolutions and with multiple references.

%% file: sections/supply.tex
\section{Supplementary material}

\subsection{SparseContext for Additional Models} 
We additionally present results of our method on two different Flux models - i) \textbf{Flux2-Klien-9B-base} - a reference conditioned image generation model trained for reference image conditioned generation and ii) \textbf{Step distilled Flux2-Klien-9B} model inferred with $8$ inference steps. The results are shown in Tab.~\ref{tab:tab-new1},~\ref{tab:tab-new2}. Our fine-tuning strategy significantly outperforms the Na\"{i}ve token dropping strategy in terms of image fidelity and text alignment while being equally fast. For the base model, our method achieves more than $2\times$ speedup when $5\%$ of the tokens are preserved. This showcases the generalization of our approach in accelerating different model architectures, including faster few-step inference methods.  

\begin{table}[h]
\caption{\textbf{Flux2-Klien-9B-base} Evaluation of generation quality with SparseContext.}
\resizebox{\textwidth}{!}{%
\begin{tabular}{@{}lccccccc|cccccc@{}}
\toprule
 & \multicolumn{7}{c|}{\textbf{Instruction Driven Image Editing}} & \multicolumn{6}{c}{\textbf{Reference Based Personalization}} \\
\midrule
 & \textbf{Tok.} $f$ & \textbf{LPIPS} $\downarrow$ & \textbf{CLIP-I} $\uparrow$ & \textbf{DINO} $\uparrow$ & \begin{tabular}[c]{@{}c@{}}\textbf{KID} $\downarrow$ \\ \small{($\times10^{-3}$)}\end{tabular} & \textbf{CLIP T-I} $\uparrow$ & \textbf{Speedup} $\uparrow$ &
\textbf{Tok.} $f$ & \textbf{CLIP-I} $\uparrow$ & \textbf{DINO} $\uparrow$ & \textbf{CLIP T-I} $\uparrow$ & \begin{tabular}[c]{@{}c@{}}\textbf{KID} $\downarrow$ \\ \small{($\times10^{-3}$)}\end{tabular} & \textbf{Speedup} $\uparrow$ \\
\midrule
\rowcolor[HTML]{D9D9D9}
Baseline & 1.00 & 0.414 & 0.827 & 0.758 & 1.3 & 0.321 & 1.00x & 1.00 & 0.727 & 0.416 & 0.284 & 27.1 & 1.00x \\ \midrule
Na\"{i}ve & 0.05 & 0.759 & 0.657 & 0.388 & 9.3 & 0.293 & 2.21x & 0.05 & 0.674 & 0.328 & 0.2830 & \textbf{24.7} & 2.21x \\
Ours      & 0.05 & \textbf{0.688} & \textbf{0.747} & \textbf{0.592} & \textbf{3.1} & \textbf{0.314} & 2.21x & 0.05 & \textbf{0.718} & \textbf{0.396} & \textbf{0.2833} & 27.2 & 2.21x \\ \midrule
Na\"{i}ve & 0.10 & 0.740 & 0.707 & 0.497 & 4.5 & 0.306 & 1.88x & 0.10 & 0.700 & 0.372 & 0.2838 & \textbf{24.7} & 1.88x \\
Ours      & 0.10 & \textbf{0.646} & \textbf{0.773} & \textbf{0.645} & \textbf{2.6} & \textbf{0.319} & 1.88x & 0.10 & \textbf{0.728} & \textbf{0.416} & \textbf{0.2839} & 27.5 & 1.88x \\ \midrule
Na\"{i}ve & 0.20 & 0.707 & 0.753 & 0.595 & 2.9 & 0.315 & 1.66x & 0.20 & 0.717 & 0.400 & \textbf{0.285} & \textbf{25.5} & 1.66x \\
Ours      & 0.20 & \textbf{0.596} & \textbf{0.793} & \textbf{0.686} & \textbf{2.3} & \textbf{0.321} & 1.66x & 0.20 & \textbf{0.735} & \textbf{0.424} & 0.284 & 29.2 & 1.66x \\ \bottomrule
\end{tabular}%
}
\label{tab:tab-new1}
\end{table}

\begin{table}[h]
\caption{\textbf{Step distilled (8-step)} inference on Flux2-Klien-9B evaluation of generation quality}
\resizebox{\textwidth}{!}{%
\begin{tabular}{@{}lccccccc|cccccc@{}}
\toprule
 & \multicolumn{7}{c|}{\textbf{Instruction Driven Image Editing}} & \multicolumn{6}{c}{\textbf{Reference Based Personalization}} \\
\midrule
 & \textbf{Tok.} $f$ & \textbf{LPIPS} $\downarrow$ & \textbf{CLIP-I} $\uparrow$ & \textbf{DINO} $\uparrow$ & \begin{tabular}[c]{@{}c@{}}\textbf{KID} $\downarrow$ \\ \small{($\times10^{-3}$)}\end{tabular} & \textbf{CLIP T-I} $\uparrow$ & \textbf{Speedup} $\uparrow$ &
\textbf{Tok.} $f$ & \textbf{CLIP-I} $\uparrow$ & \textbf{DINO} $\uparrow$ & \textbf{CLIP T-I} $\uparrow$ & \begin{tabular}[c]{@{}c@{}}\textbf{KID} $\downarrow$ \\ \small{($\times10^{-3}$)}\end{tabular} & \textbf{Speedup} $\uparrow$ \\
\midrule
\rowcolor[HTML]{D9D9D9}
Baseline & 1.00 & 0.409 & 0.831 & 0.765 & 1.23 & 0.320 & 1.00x & 1.00 & 0.718 & 0.420 & 0.286 & 28.05 & 1.00x \\ \midrule
Na\"{i}ve & 0.05 & 0.742 & 0.681 & 0.454 & \textbf{4.3} & 0.298 & 1.55x & 0.05 & 0.662 & 0.326 & \textbf{0.287} & 28.10 & 1.55x \\
Ours      & 0.05 & \textbf{0.662} & \textbf{0.770} & \textbf{0.666} & 5.1 & \textbf{0.306} & 1.55x & 0.05 & \textbf{0.726} & \textbf{0.442} & 0.272 & \textbf{25.23} & 1.55x \\ \midrule
Na\"{i}ve & 0.10 & 0.717 & 0.733 & 0.565 & \textbf{3.1} & 0.309 & 1.56x & 0.10 & 0.690 & 0.370 & \textbf{0.287} & 27.13 & 1.56x \\
Ours      & 0.10 & \textbf{0.628} & \textbf{0.791} & \textbf{0.706} & 4.1 & \textbf{0.313} & 1.56x & 0.10 & \textbf{0.736} & \textbf{0.461} & 0.272 & \textbf{24.28} & 1.56x \\ \midrule
Na\"{i}ve & 0.20 & 0.680 & 0.772 & 0.645 & \textbf{2.5} & \textbf{0.316} & 1.43x & 0.20 & 0.704 & 0.399 & \textbf{0.287} & 27.00 & 1.43x \\
Ours      & 0.20 & \textbf{0.576} & \textbf{0.809} & \textbf{0.740} & 3.0 & 0.316 & 1.43x & 0.20 & \textbf{0.743} & \textbf{0.464} & 0.272 & \textbf{24.79} & 1.43x \\ \bottomrule
\end{tabular}%
}
\label{tab:tab-new2}
\end{table}

\subsection{Additional Results}
\begin{figure}[h]
    \centering
    \includegraphics[width=1.0\linewidth]{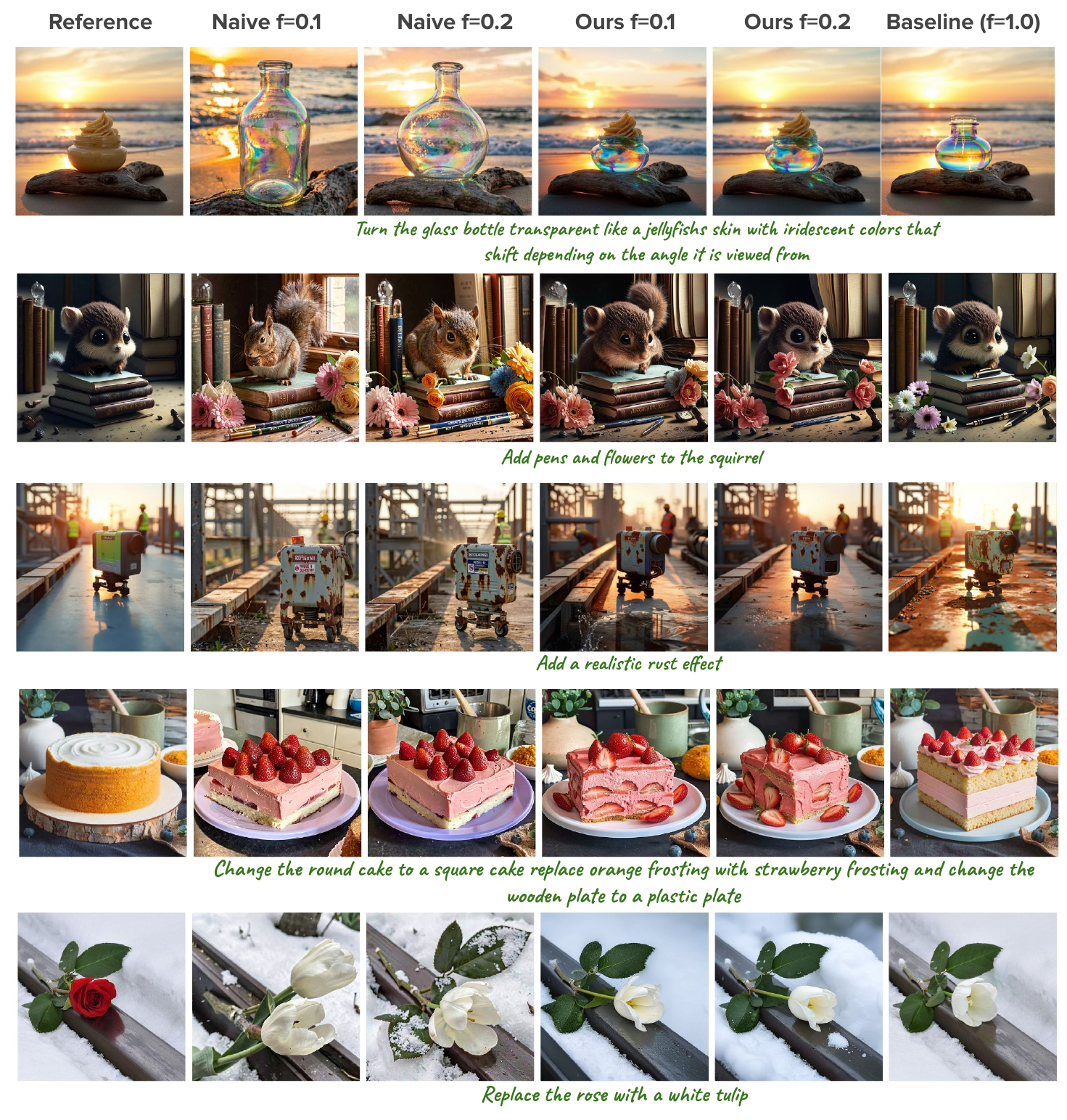}
    \caption{Results for efficient instruction-based image editing.}
    \label{fig:addn-editing-results}
\end{figure}

\begin{figure}[h]
    \centering
    \includegraphics[width=1.0\linewidth]{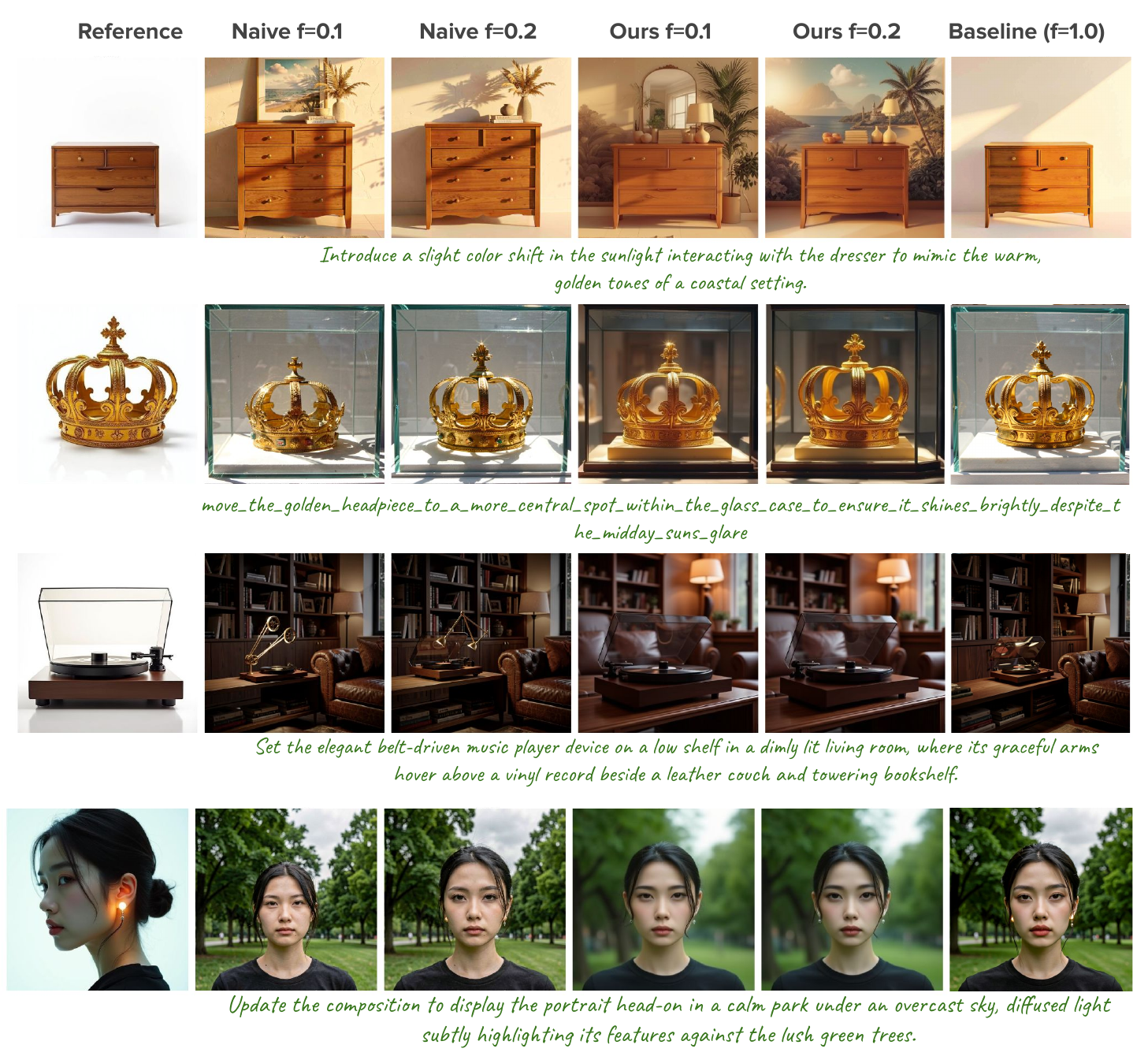}
    \caption{Results for single-reference personalization with \methodname{}.}
    \label{fig:single-reference-personalization}
\end{figure}

\begin{figure}[h]
    \centering
    \includegraphics[width=1.0\linewidth]{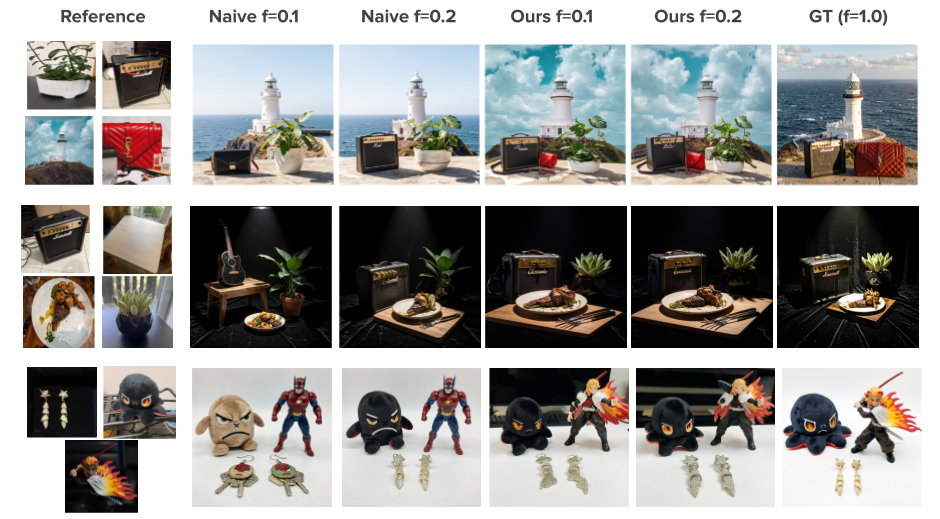}
    \caption{Results for multi-reference personalization with \methodname{}. We compare the Na\"{i}ve baseline and \methodname{} at token drop fractions $f=0.1$ and $f=0.2$.}
    \label{fig:multi-reference-personalization}
\end{figure}

We present additional qualitative results for \methodname{} against the Na\"{i}ve baseline for instruction-driven image editing in Fig.~\ref{fig:addn-editing-results}, for single-reference personalization in Fig.~\ref{fig:single-reference-personalization}, and for multi-reference personalization in Fig.~\ref{fig:multi-reference-personalization}.

\subsection{Implementation Details.}
\begin{figure}
    \centering
    \includegraphics[width=0.8\linewidth]{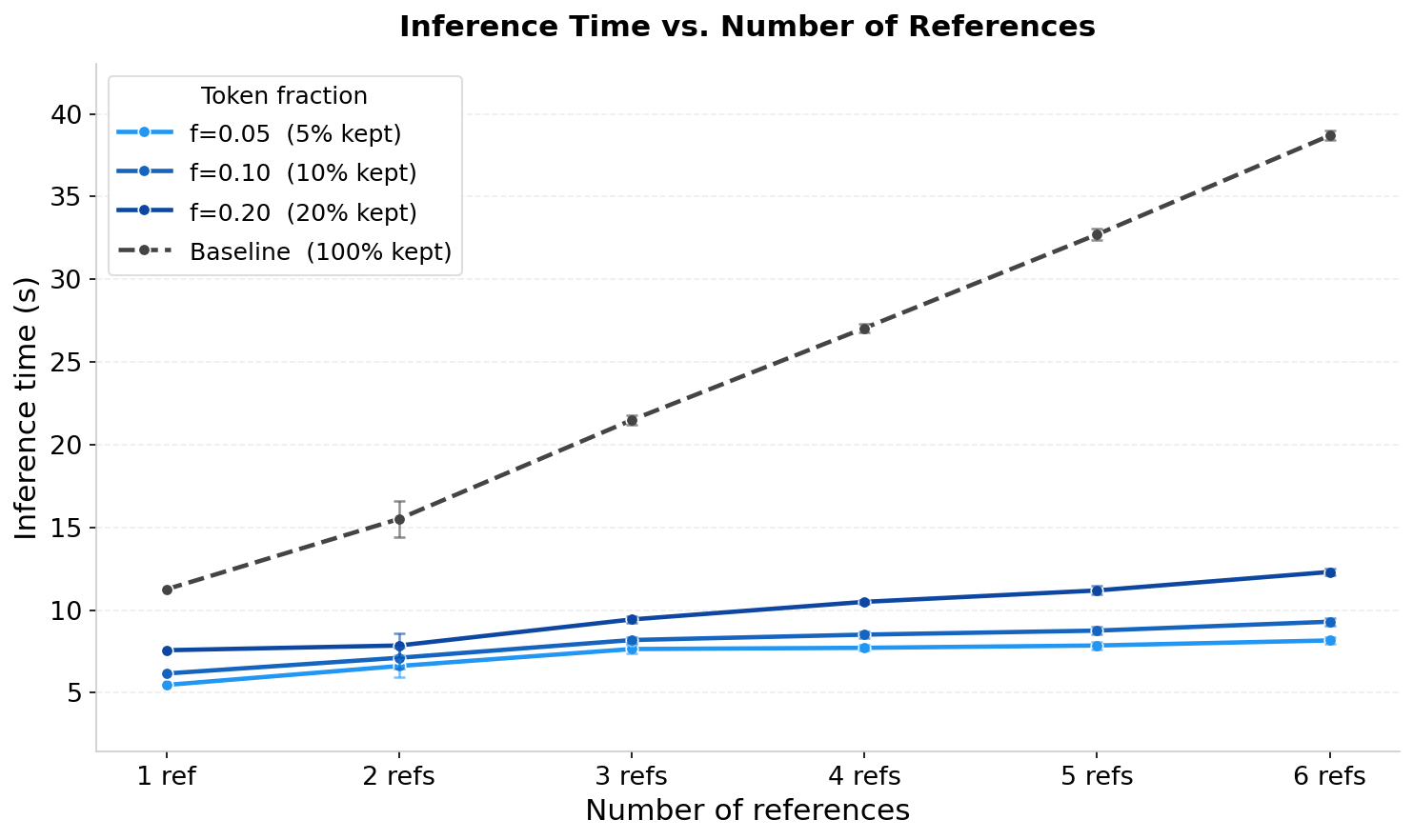}
    \caption{Clock time for inference on A100 GPU}
    \label{fig:true-time}
\end{figure}

We use FLUX.2 Klien-9B~\cite{flux-2-2025} model for all our experiments. We used $28$ step schedule for training and inference of the model. The training took $20$ hrs on two A100 GPUS. The learning rate was $1e-4$ with a constant learning warmup and the training use $4$ steps of gradient accumulation. The LoRA rank is $16$ and is learned only over the attention layer parameters. For inference tasks, we use first apply a Gaussian filter of $3\times3$ on the input image and then extract Canny edge map for image editing application and use off-the-shelf saliency prediction network~\cite{kummerer2025modeling} to obtain a single channel saliency map for the reference images. We additionally present the true runtime on A100 machine in the Fig.~\ref{fig:true-time}

\subsection{Dataset Generation}
\noindent

\paragraph{Instruction-driven editing dataset.}

\noindent 
\textbf{Training.} For training we use a subset of HQ-edit
edit dataset. We used the source image and edit instructions from HQ-edit and then generate their corresponding edited images using Flux-Klien-9B model to obtain the ground truth edits, as the targets provided in the original dataset are not accurate.  

\textbf{Evaluation.}
We evaluate instruction-driven image editing on the PIE-Bench dataset. Since PIE-Bench provides source and target prompt pairs for each edit, we use Qwen3-8B to automatically generate natural language editing instructions from each pair, yielding a set of $(\texttt{source image},\ \texttt{edit instruction})$ pairs used throughout evaluation.

\textbf{Single reference personalization.}
For both training and testing of single reference personalization, we use a subset of Subject-200K dataset consisting of simple objects in different compositions. We selected a 30K dataset for training and another 500 samples for evaluation.

\vspace{0.5em}
\noindent
\paragraph{Multi-reference personalization dataset.}
For multi-subject personalization, we construct a synthetic compositional dataset using objects from the Custom Diffusion Objects dataset, grouping object instances into semantic categories solely for structured sampling. We generate approximately $100$ prompt templates covering varying object counts and spatial relations, e.g., \textit{``Place \{a\} on top of \{b\} while \{c\} is placed beside \{d\}''}, where placeholders are populated using randomly sampled object instances from the categorized groups. For each example, the number of objects is drawn from a categorical distribution over $\{2,3,4,5,6\}$ with probabilities $[0.3, 0.3, 0.2, 0.1, 0.1]$, respectively. A prompt template corresponding to the sampled object count is then selected and instantiated accordingly. This procedure yields a diverse set of multi-object compositions with varied spatial arrangements and interaction patterns.

\subsection{Personalization with Multi-Reference Images}

\begin{table}[h]
\centering
\caption{CLIP text-to-image similarity for multi-reference personalization using saliency-based conditioning across different token drop fractions.}
\label{tab:multi-ref-personalization}
\begin{tabular}{lcc}
\toprule
\textbf{Method} & \textbf{Token \%} & \textbf{CLIP T-I} \\
\midrule
Baseline & 100\% & 0.347 \\
\midrule
Naïve   & 5\%  & 0.361 \\
Ours    & 5\%  & 0.341 \\
\midrule
Naïve   & 10\% & 0.359 \\
Ours    & 10\% & 0.337 \\
\midrule
Naïve   & 20\% & 0.353 \\
Ours    & 20\% & 0.332 \\
\bottomrule
\end{tabular}
\end{table}



\subsection{Memory overhead}
\begin{figure}[h]
    \centering
    \includegraphics[width=1.0\linewidth]{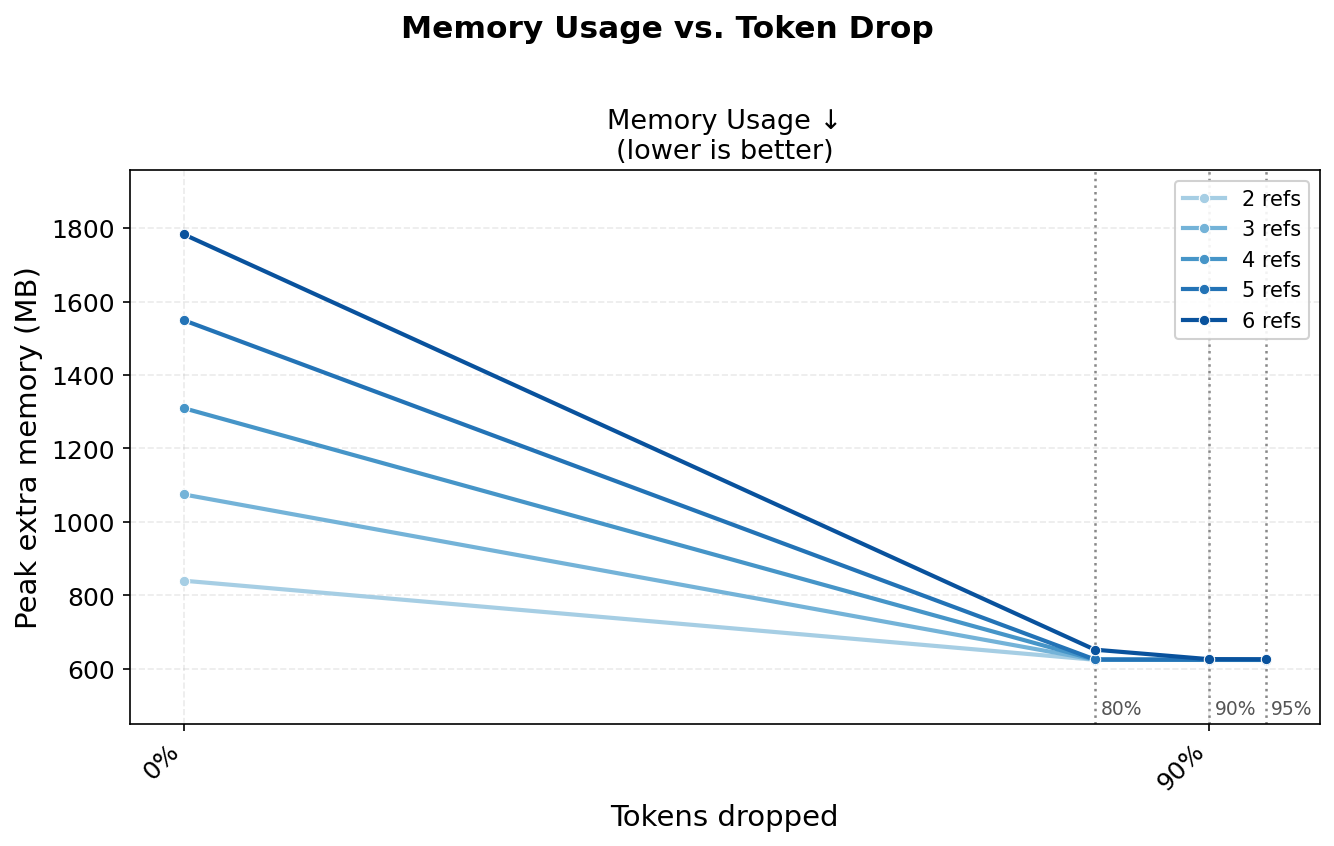}
    \caption{Peak memory during inference with number of reference image and differen token dropping fractions}
    \label{fig:memory-plot}
\end{figure}

We benchmark the additional dynamic memory overhead during the inference process in Fig.~\ref{fig:memory-plot}. As we scale the number of reference images, our method significantly reduced memory consumptions as compared to the baseline without token dropping. This experiment was performed on RTX-6000 Pro with 96G to fit the larger image resolution.

\subsection{Limitation}
\begin{figure}
    \centering
    \includegraphics[width=0.9\linewidth]{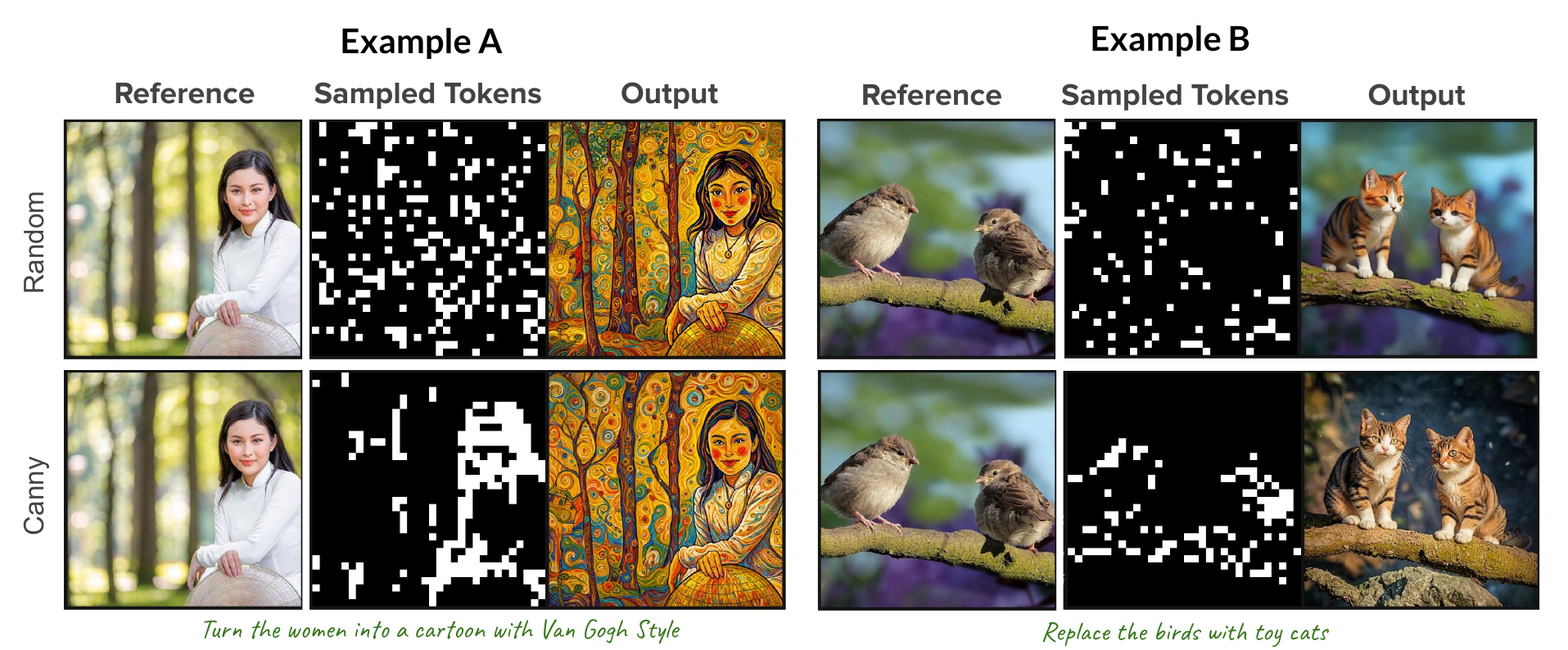}
    \caption{\textbf{Limitation.} The choice of the inference time token selection heuristic can depend on the type of edit. In example A stylization edit Canny based sampling is effective but for example B, it will focus only the object region that needs to be replaced, resulting in inferior quality of the background region.}
    \label{fig:limitation}
\end{figure}

Though \methodname{} enables inference time control over selecting a best suited approach selecting tokens, the selection process may depend on the input image. Consider the scenario given in Fig.~\ref{fig:limitation} about image editing, where we show two different editing scenarios. We use canny edge based token sampling for editing. In example A, the edit is stylization and using Canny edge preserves the girls identity well as compared to random selection. However in example B, the edit instruction is to replace the object, and using canny based sampling can hurt the preservation of background region as the structure boundary of the bird is dominant. 

%% file: main.bib
@String(CVPR= {IEEE Conf. Comput. Vis. Pattern Recog.})

@String(CVPR  = {CVPR})

@article{wu2025qwen,
  title={Qwen-image technical report},
  author={Wu, Chenfei and Li, Jiahao and Zhou, Jingren and Lin, Junyang and Gao, Kaiyuan and Yan, Kun and Yin, Sheng-ming and Bai, Shuai and Xu, Xiao and Chen, Yilei and others},
  journal={arXiv preprint arXiv:2508.02324},
  year={2025}
}

@inproceedings{zhang2018unreasonable,
  title={The unreasonable effectiveness of deep features as a perceptual metric},
  author={Zhang, Richard and Isola, Phillip and Efros, Alexei A and Shechtman, Eli and Wang, Oliver},
  booktitle={Proceedings of the IEEE conference on computer vision and pattern recognition},
  pages={586--595},
  year={2018}
}

@article{Qwen2.5-VL,
  title={Qwen2.5-VL Technical Report},
  author={Bai, Shuai and Chen, Keqin and Liu, Xuejing and Wang, Jialin and Ge, Wenbin and Song, Sibo and Dang, Kai and Wang, Peng and Wang, Shijie and Tang, Jun and Zhong, Humen and Zhu, Yuanzhi and Yang, Mingkun and Li, Zhaohai and Wan, Jianqiang and Wang, Pengfei and Ding, Wei and Fu, Zheren and Xu, Yiheng and Ye, Jiabo and Zhang, Xi and Xie, Tianbao and Cheng, Zesen and Zhang, Hang and Yang, Zhibo and Xu, Haiyang and Lin, Junyang},
  journal={arXiv preprint arXiv:2502.13923},
  year={2025}
}

@inproceedings{brooks2023instructpix2pix,
  title={Instructpix2pix: Learning to follow image editing instructions},
  author={Brooks, Tim and Holynski, Aleksander and Efros, Alexei A},
  booktitle={Proceedings of the IEEE/CVF conference on computer vision and pattern recognition},
  pages={18392--18402},
  year={2023}
}

@inproceedings{sheynin2024emu,
  title={Emu edit: Precise image editing via recognition and generation tasks},
  author={Sheynin, Shelly and Polyak, Adam and Singer, Uriel and Kirstain, Yuval and Zohar, Amit and Ashual, Oron and Parikh, Devi and Taigman, Yaniv},
  booktitle={Proceedings of the IEEE/CVF Conference on Computer Vision and Pattern Recognition},
  pages={8871--8879},
  year={2024}
}

@article{batifol2025flux,
  title={FLUX. 1 Kontext: Flow Matching for In-Context Image Generation and Editing in Latent Space},
  author={Batifol, Stephen and Blattmann, Andreas and Boesel, Frederic and Consul, Saksham and Diagne, Cyril and Dockhorn, Tim and English, Jack and English, Zion and Esser, Patrick and Kulal, Sumith and others},
  journal={arXiv e-prints},
  pages={arXiv--2506},
  year={2025}
}

@inproceedings{parihar2024precisecontrol,
  title={Precisecontrol: Enhancing text-to-image diffusion models with fine-grained attribute control},
  author={Parihar, Rishubh and Sachidanand, VS and Mani, Sabariswaran and Karmali, Tejan and Venkatesh Babu, R},
  booktitle={European Conference on Computer Vision},
  pages={469--487},
  year={2024},
  organization={Springer}
}

@inproceedings{peebles2023scalable,
  title={Scalable diffusion models with transformers},
  author={Peebles, William and Xie, Saining},
  booktitle={Proceedings of the IEEE/CVF international conference on computer vision},
  pages={4195--4205},
  year={2023}
}

@misc{song2022denoisingdiffusionimplicitmodels,
      title={Denoising Diffusion Implicit Models}, 
      author={Jiaming Song and Chenlin Meng and Stefano Ermon},
      year={2022},
      eprint={2010.02502},
      archivePrefix={arXiv},
      primaryClass={cs.LG},
      url={https://arxiv.org/abs/2010.02502}, 
}

@misc{oquab2023dinov2,
  title={DINOv2: Learning Robust Visual Features without Supervision},
  author={Oquab, Maxime and Darcet, Timothée and Moutakanni, Theo and Vo, Huy V. and Szafraniec, Marc and Khalidov, Vasil and Fernandez, Pierre and Haziza, Daniel and Massa, Francisco and El-Nouby, Alaaeldin and Howes, Russell and Huang, Po-Yao and Xu, Hu and Sharma, Vasu and Li, Shang-Wen and Galuba, Wojciech and Rabbat, Mike and Assran, Mido and Ballas, Nicolas and Synnaeve, Gabriel and Misra, Ishan and Jegou, Herve and Mairal, Julien and Labatut, Patrick and Joulin, Armand and Bojanowski, Piotr},
  journal={arXiv:2304.07193},
  year={2023}
}

@InProceedings{Rombach_2022_CVPR,
    author    = {Rombach, Robin and Blattmann, Andreas and Lorenz, Dominik and Esser, Patrick and Ommer, Bj\"orn},
    title     = {High-Resolution Image Synthesis With Latent Diffusion Models},
    booktitle = {Proceedings of the IEEE/CVF Conference on Computer Vision and Pattern Recognition (CVPR)},
    month     = {June},
    year      = {2022},
    pages     = {10684-10695}
}

@misc{saharia2022photorealistictexttoimagediffusionmodels,
      title={Photorealistic Text-to-Image Diffusion Models with Deep Language Understanding}, 
      author={Chitwan Saharia and William Chan and Saurabh Saxena and Lala Li and Jay Whang and Emily Denton and Seyed Kamyar Seyed Ghasemipour and Burcu Karagol Ayan and S. Sara Mahdavi and Rapha Gontijo Lopes and Tim Salimans and Jonathan Ho and David J Fleet and Mohammad Norouzi},
      year={2022},
      eprint={2205.11487},
      archivePrefix={arXiv},
      primaryClass={cs.CV},
      url={https://arxiv.org/abs/2205.11487}, 
}

@misc{dhariwal2021diffusionmodelsbeatgans,
      title={Diffusion Models Beat GANs on Image Synthesis}, 
      author={Prafulla Dhariwal and Alex Nichol},
      year={2021},
      eprint={2105.05233},
      archivePrefix={arXiv},
      primaryClass={cs.LG},
      url={https://arxiv.org/abs/2105.05233}, 
}

@article{tan2024ominicontrol,
  title={Ominicontrol: Minimal and universal control for diffusion transformer},
  author={Tan, Zhenxiong and Liu, Songhua and Yang, Xingyi and Xue, Qiaochu and Wang, Xinchao},
  journal={arXiv preprint arXiv:2411.15098},
  year={2024}
}

@inproceedings{huihq,
  title={HQ-Edit: A High-Quality Dataset for Instruction-based Image Editing},
  author={Hui, Mude and Yang, Siwei and Zhao, Bingchen and Shi, Yichun and Wang, Heng and Wang, Peng and Xie, Cihang and Zhou, Yuyin},
  booktitle={The Thirteenth International Conference on Learning Representations}
}

@misc{ye2023ipadaptertextcompatibleimage,
      title={IP-Adapter: Text Compatible Image Prompt Adapter for Text-to-Image Diffusion Models}, 
      author={Hu Ye and Jun Zhang and Sibo Liu and Xiao Han and Wei Yang},
      year={2023},
      eprint={2308.06721},
      archivePrefix={arXiv},
      primaryClass={cs.CV},
      url={https://arxiv.org/abs/2308.06721}, 
}

@article{hertz2022prompt,
  title={Prompt-to-prompt image editing with cross attention control},
  author={Hertz, Amir and Mokady, Ron and Tenenbaum, Jay and Aberman, Kfir and Pritch, Yael and Cohen-Or, Daniel},
  booktitle={arXiv preprint arXiv:2208.01626},
  year={2022}
}

@InProceedings{Mokady_2023_CVPR,
    author    = {Mokady, Ron and Hertz, Amir and Aberman, Kfir and Pritch, Yael and Cohen-Or, Daniel},
    title     = {NULL-Text Inversion for Editing Real Images Using Guided Diffusion Models},
    booktitle = {Proceedings of the IEEE/CVF Conference on Computer Vision and Pattern Recognition (CVPR)},
    month     = {June},
    year      = {2023},
    pages     = {6038-6047}
}

@misc{meng2022sdeditguidedimagesynthesis,
      title={SDEdit: Guided Image Synthesis and Editing with Stochastic Differential Equations}, 
      author={Chenlin Meng and Yutong He and Yang Song and Jiaming Song and Jiajun Wu and Jun-Yan Zhu and Stefano Ermon},
      year={2022},
      eprint={2108.01073},
      archivePrefix={arXiv},
      primaryClass={cs.CV},
      url={https://arxiv.org/abs/2108.01073}, 
}

@misc{alaluf2023crossimageattentionzeroshotappearance,
      title={Cross-Image Attention for Zero-Shot Appearance Transfer}, 
      author={Yuval Alaluf and Daniel Garibi and Or Patashnik and Hadar Averbuch-Elor and Daniel Cohen-Or},
      year={2023},
      eprint={2311.03335},
      archivePrefix={arXiv},
      primaryClass={cs.CV},
      url={https://arxiv.org/abs/2311.03335}, 
}

@InProceedings{guo2024pulid,
  title={PuLID: Pure and Lightning ID Customization via Contrastive Alignment},
  author={Guo, Zinan and Wu, Yanze and Chen, Zhuowei and Chen, Lang and Zhang, Peng and He, Qian},
  booktitle={Advances in Neural Information Processing Systems},
  year={2024}
}

@misc{gal2024lcmlookahead,
    title={LCM-Lookahead for Encoder-based Text-to-Image Personalization}, 
    author={Rinon Gal and Or Lichter and Elad Richardson and Or Patashnik and Amit H. Bermano and Gal Chechik and Daniel Cohen-Or},
    year={2024},
    eprint={2404.03620},
    archivePrefix={arXiv},
    primaryClass={cs.CV}
}

@inproceedings{ruiz2023dreambooth,
  title={Dreambooth: Fine tuning text-to-image diffusion models for subject-driven generation},
  author={Ruiz, Nataniel and Li, Yuanzhen and Jampani, Varun and Pritch, Yael and Rubinstein, Michael and Aberman, Kfir},
  booktitle={Proceedings of the IEEE/CVF conference on computer vision and pattern recognition},
  pages={22500--22510},
  year={2023}
}

@inproceedings{zeng2024jedi,
                title={JeDi: Joint-image Diffusion Models for Finetuning-free Personalized Text-to-image Generation},
                author={Zeng, Yu and Patel, Vishal M and Wang, Haochen and Huang, Xun and Wang, Ting-Chun and Liu, Ming-Yu and Balaji, Yogesh},
                booktitle={Proceedings of the IEEE/CVF Conference on Computer Vision and Pattern Recognition},
                year={2024}
              }

@misc{tumanyan2022plugandplaydiffusionfeaturestextdriven,
      title={Plug-and-Play Diffusion Features for Text-Driven Image-to-Image Translation}, 
      author={Narek Tumanyan and Michal Geyer and Shai Bagon and Tali Dekel},
      year={2022},
      eprint={2211.12572},
      archivePrefix={arXiv},
      primaryClass={cs.CV},
      url={https://arxiv.org/abs/2211.12572}, 
}

@misc{hubermanspiegelglas2024editfriendlyddpmnoise,
      title={An Edit Friendly DDPM Noise Space: Inversion and Manipulations}, 
      author={Inbar Huberman-Spiegelglas and Vladimir Kulikov and Tomer Michaeli},
      year={2024},
      eprint={2304.06140},
      archivePrefix={arXiv},
      primaryClass={cs.CV},
      url={https://arxiv.org/abs/2304.06140}, 
}

@misc{gal2022imageworthwordpersonalizing,
      title={An Image is Worth One Word: Personalizing Text-to-Image Generation using Textual Inversion}, 
      author={Rinon Gal and Yuval Alaluf and Yuval Atzmon and Or Patashnik and Amit H. Bermano and Gal Chechik and Daniel Cohen-Or},
      year={2022},
      eprint={2208.01618},
      archivePrefix={arXiv},
      primaryClass={cs.CV},
      url={https://arxiv.org/abs/2208.01618}, 
}

@misc{cao2023masactrltuningfreemutualselfattention,
      title={MasaCtrl: Tuning-Free Mutual Self-Attention Control for Consistent Image Synthesis and Editing}, 
      author={Mingdeng Cao and Xintao Wang and Zhongang Qi and Ying Shan and Xiaohu Qie and Yinqiang Zheng},
      year={2023},
      eprint={2304.08465},
      archivePrefix={arXiv},
      primaryClass={cs.CV},
      url={https://arxiv.org/abs/2304.08465}, 
}

@misc{xlabs-flux-ip-adapter,
  title = {XLabs-AI FLUX Ip-Adapter},
  howpublished = {\url{https://huggingface.co/XLabs-AI/flux-ip-adapter}},
  year={2024}
}

@inproceedings{lipmanflow,
  title={Flow Matching for Generative Modeling},
  author={Lipman, Yaron and Chen, Ricky TQ and Ben-Hamu, Heli and Nickel, Maximilian and Le, Matthew},
  booktitle={The Eleventh International Conference on Learning Representations},
  year={2023}
}

@misc{radford2021learning,
    title={Learning Transferable Visual Models From Natural Language Supervision},
    author={Alec Radford and Jong Wook Kim and Chris Hallacy and Aditya Ramesh and Gabriel Goh and Sandhini Agarwal and Girish Sastry and Amanda Askell and Pamela Mishkin and Jack Clark and Gretchen Krueger and Ilya Sutskever},
    year={2021},
    eprint={2103.00020},
    archivePrefix={arXiv},
    primaryClass={cs.CV}
}

@article{10.1109/TPAMI.2025.3541625,
author = {Huang, Yi and Huang, Jiancheng and Liu, Yifan and Yan, Mingfu and Lv, Jiaxi and Liu, Jianzhuang and Xiong, Wei and Zhang, He and Cao, Liangliang and Chen, Shifeng},
title = {Diffusion Model-Based Image Editing: A Survey},
year = {2025},
issue_date = {June 2025},
publisher = {IEEE Computer Society},
address = {USA},
volume = {47},
number = {6},
issn = {0162-8828},
url = {https://doi.org/10.1109/TPAMI.2025.3541625},
doi = {10.1109/TPAMI.2025.3541625},
journal = {IEEE Trans. Pattern Anal. Mach. Intell.},
month = jun,
pages = {4409–4437},
numpages = {29}
}

@inproceedings{kulikov2025flowedit,
  title={Flowedit: Inversion-free text-based editing using pre-trained flow models},
  author={Kulikov, Vladimir and Kleiner, Matan and Huberman-Spiegelglas, Inbar and Michaeli, Tomer},
  booktitle={Proceedings of the IEEE/CVF International Conference on Computer Vision},
  pages={19721--19730},
  year={2025}
}

@inproceedings{kumari2023multi,
  title={Multi-concept customization of text-to-image diffusion},
  author={Kumari, Nupur and Zhang, Bingliang and Zhang, Richard and Shechtman, Eli and Zhu, Jun-Yan},
  booktitle={Proceedings of the IEEE/CVF conference on computer vision and pattern recognition},
  pages={1931--1941},
  year={2023}
}

@inproceedings{wei2023elite,
  title={Elite: Encoding visual concepts into textual embeddings for customized text-to-image generation},
  author={Wei, Yuxiang and Zhang, Yabo and Ji, Zhilong and Bai, Jinfeng and Zhang, Lei and Zuo, Wangmeng},
  booktitle={Proceedings of the IEEE/CVF international conference on computer vision},
  pages={15943--15953},
  year={2023}
}

@article{https://doi.org/10.1111/cgf.15063,
author = {Po, R. and Yifan, W. and Golyanik, V. and Aberman, K. and Barron, J. T. and Bermano, A. and Chan, E. and Dekel, T. and Holynski, A. and Kanazawa, A. and Liu, C.K. and Liu, L. and Mildenhall, B. and Nießner, M. and Ommer, B. and Theobalt, C. and Wonka, P. and Wetzstein, G.},
title = {State of the Art on Diffusion Models for Visual Computing},
journal = {Computer Graphics Forum},
volume = {43},
number = {2},
pages = {e15063},
doi = {https://doi.org/10.1111/cgf.15063},
url = {https://onlinelibrary.wiley.com/doi/abs/10.1111/cgf.15063},
eprint = {https://onlinelibrary.wiley.com/doi/pdf/10.1111/cgf.15063},
year = {2024}
}

@misc{flux,
  key = {https://github.com/black-forest-labs/flux},
  author = {Black Forest Labs},
  title = {FLUX},
  year = {2024},
  howpublished = {\url{https://github.com/black-forest-labs/flux}},
}

@inproceedings{Parmar_2023, series={SIGGRAPH ’23},
   title={Zero-shot Image-to-Image Translation},
   url={http://dx.doi.org/10.1145/3588432.3591513},
   DOI={10.1145/3588432.3591513},
   booktitle={Special Interest Group on Computer Graphics and Interactive Techniques Conference Conference Proceedings},
   publisher={ACM},
   author={Parmar, Gaurav and Kumar Singh, Krishna and Zhang, Richard and Li, Yijun and Lu, Jingwan and Zhu, Jun-Yan},
   year={2023},
   month=jul, pages={1–11},
   collection={SIGGRAPH ’23} }

@article{garibi2024renoise,
  title={ReNoise: Real Image Inversion Through Iterative Noising},
  author={Garibi, Daniel and Patashnik, Or and Voynov, Andrey and Averbuch-Elor, Hadar and Cohen-Or, Daniel},
  journal={arXiv preprint arXiv:2403.14602},
  year={2024}
}

@inproceedings{shin2025large,
  title={Large-scale text-to-image model with inpainting is a zero-shot subject-driven image generator},
  author={Shin, Chaehun and Choi, Jooyoung and Kim, Heeseung and Yoon, Sungroh},
  booktitle={Proceedings of the Computer Vision and Pattern Recognition Conference},
  pages={7986--7996},
  year={2025}
}

@inproceedings{avrahami2025stable,
  title={Stable flow: Vital layers for training-free image editing},
  author={Avrahami, Omri and Patashnik, Or and Fried, Ohad and Nemchinov, Egor and Aberman, Kfir and Lischinski, Dani and Cohen-Or, Daniel},
  booktitle={Proceedings of the Computer Vision and Pattern Recognition Conference},
  pages={7877--7888},
  year={2025}
}

@inproceedings{kumari2025generating,
  title={Generating multi-image synthetic data for text-to-image customization},
  author={Kumari, Nupur and Yin, Xi and Zhu, Jun-Yan and Misra, Ishan and Azadi, Samaneh},
  booktitle={Proceedings of the IEEE/CVF International Conference on Computer Vision},
  pages={16524--16534},
  year={2025}
}

@inproceedings{cai2025diffusion,
  title={Diffusion self-distillation for zero-shot customized image generation},
  author={Cai, Shengqu and Chan, Eric Ryan and Zhang, Yunzhi and Guibas, Leonidas and Wu, Jiajun and Wetzstein, Gordon},
  booktitle={Proceedings of the Computer Vision and Pattern Recognition Conference},
  pages={18434--18443},
  year={2025}
}

@misc{flux-2-2025,
    author={Black Forest Labs},
    title={{FLUX.2: Frontier Visual Intelligence}},
    year={2025},
    howpublished={\url{https://bfl.ai/blog/flux-2}},
}

@online{deepmind2025nanobanana,
  author    = {{Google DeepMind}},
  title     = {Gemini Image - Nano Banana},
  year      = {2025},
  url       = {https://deepmind.google/models/gemini-image/},
  urldate   = {2026-04-28}
}

@article{beltagy2020longformer,
  title={Longformer: The long-document transformer},
  author={Beltagy, Iz and Peters, Matthew E and Cohan, Arman},
  journal={arXiv preprint arXiv:2004.05150},
  year={2020}
}

@article{choromanski2020rethinking,
  title={Rethinking attention with performers},
  author={Choromanski, Krzysztof and Likhosherstov, Valerii and Dohan, David and Song, Xingyou and Gane, Andreea and Sarlos, Tamas and Hawkins, Peter and Davis, Jared and Mohiuddin, Afroz and Kaiser, Lukasz and others},
  journal={arXiv preprint arXiv:2009.14794},
  year={2020}
}

@inproceedings{katharopoulos2020transformers,
  title={Transformers are rnns: Fast autoregressive transformers with linear attention},
  author={Katharopoulos, Angelos and Vyas, Apoorv and Pappas, Nikolaos and Fleuret, Fran{\c{c}}ois},
  booktitle={International conference on machine learning},
  pages={5156--5165},
  year={2020},
  organization={PMLR}
}

@inproceedings{shen2021efficient,
  title={Efficient attention: Attention with linear complexities},
  author={Shen, Zhuoran and Zhang, Mingyuan and Zhao, Haiyu and Yi, Shuai and Li, Hongsheng},
  booktitle={Proceedings of the IEEE/CVF winter conference on applications of computer vision},
  pages={3531--3539},
  year={2021}
}

@inproceedings{han2023flatten,
  title={Flatten transformer: Vision transformer using focused linear attention},
  author={Han, Dongchen and Pan, Xuran and Han, Yizeng and Song, Shiji and Huang, Gao},
  booktitle={Proceedings of the IEEE/CVF international conference on computer vision},
  pages={5961--5971},
  year={2023}
}

@article{dao2022flashattention,
  title={Flashattention: Fast and memory-efficient exact attention with io-awareness},
  author={Dao, Tri and Fu, Dan and Ermon, Stefano and Rudra, Atri and R{\'e}, Christopher},
  journal={Advances in neural information processing systems},
  volume={35},
  pages={16344--16359},
  year={2022}
}

@article{wang2020linformer,
  title={Linformer: Self-attention with linear complexity},
  author={Wang, Sinong and Li, Belinda Z and Khabsa, Madian and Fang, Han and Ma, Hao},
  journal={arXiv preprint arXiv:2006.04768},
  year={2020}
}

@inproceedings{xia2025training,
  title={Training-free and adaptive sparse attention for efficient long video generation},
  author={Xia, Yifei and Ling, Suhan and Fu, Fangcheng and Wang, Yujie and Li, Huixia and Xiao, Xuefeng and Cui, Bin},
  booktitle={Proceedings of the IEEE/CVF International Conference on Computer Vision},
  pages={15982--15993},
  year={2025}
}

@inproceedings{ma2024deepcache,
  title={Deepcache: Accelerating diffusion models for free},
  author={Ma, Xinyin and Fang, Gongfan and Wang, Xinchao},
  booktitle={Proceedings of the IEEE/CVF conference on computer vision and pattern recognition},
  pages={15762--15772},
  year={2024}
}

@inproceedings{wimbauer2024cache,
  title={Cache me if you can: Accelerating diffusion models through block caching},
  author={Wimbauer, Felix and Wu, Bichen and Schoenfeld, Edgar and Dai, Xiaoliang and Hou, Ji and He, Zijian and Sanakoyeu, Artsiom and Zhang, Peizhao and Tsai, Sam and Kohler, Jonas and others},
  booktitle={Proceedings of the IEEE/CVF Conference on Computer Vision and Pattern Recognition},
  pages={6211--6220},
  year={2024}
}

@article{ma2024learning,
  title={Learning-to-cache: Accelerating diffusion transformer via layer caching},
  author={Ma, Xinyin and Fang, Gongfan and Bi Mi, Michael and Wang, Xinchao},
  journal={Advances in Neural Information Processing Systems},
  volume={37},
  pages={133282--133304},
  year={2024}
}

@inproceedings{liu2025timestep,
  title={Timestep Embedding Tells: It's Time to Cache for Video Diffusion Model},
  author={Liu, Feng and Zhang, Shiwei and Wang, Xiaofeng and Wei, Yujie and Qiu, Haonan and Zhao, Yuzhong and Zhang, Yingya and Ye, Qixiang and Wan, Fang},
  booktitle={Proceedings of the Computer Vision and Pattern Recognition Conference},
  pages={7353--7363},
  year={2025}
}

@inproceedings{kahatapitiya2025adaptive,
  title={Adaptive caching for faster video generation with diffusion transformers},
  author={Kahatapitiya, Kumara and Liu, Haozhe and He, Sen and Liu, Ding and Jia, Menglin and Zhang, Chenyang and Ryoo, Michael S and Xie, Tian},
  booktitle={Proceedings of the IEEE/CVF International Conference on Computer Vision},
  pages={15240--15252},
  year={2025}
}

@article{hu2025flashdlm,
  title={FlashDLM: Accelerating Diffusion Language Model Inference via Efficient KV Caching and Guided Diffusion},
  author={Hu, Zhanqiu and Meng, Jian and Akhauri, Yash and Abdelfattah, Mohamed S and Seo, Jae-sun and Zhang, Zhiru and Gupta, Udit},
  journal={arXiv preprint arXiv:2505.21467},
  year={2025}
}

@inproceedings{bolyatoken,
  title={Token Merging: Your ViT But Faster},
  author={Bolya, Daniel and Fu, Cheng-Yang and Dai, Xiaoliang and Zhang, Peizhao and Feichtenhofer, Christoph and Hoffman, Judy},
  booktitle={The Eleventh International Conference on Learning Representations}
}

@inproceedings{kim2024token,
  title={Token fusion: Bridging the gap between token pruning and token merging},
  author={Kim, Minchul and Gao, Shangqian and Hsu, Yen-Chang and Shen, Yilin and Jin, Hongxia},
  booktitle={Proceedings of the IEEE/CVF Winter Conference on Applications of Computer Vision},
  pages={1383--1392},
  year={2024}
}

@inproceedings{bolya2023token,
  title={Token merging for fast stable diffusion},
  author={Bolya, Daniel and Hoffman, Judy},
  booktitle={Proceedings of the IEEE/CVF conference on computer vision and pattern recognition},
  pages={4599--4603},
  year={2023}
}

@inproceedings{lu2025toma,
  title={ToMA: Token Merge with Attention for Diffusion Models},
  author={Lu, Wenbo and Zheng, Shaoyi and Xia, Yuxuan and Wang, Shengjie},
  booktitle={International Conference on Machine Learning},
  pages={40930--40951},
  year={2025},
  organization={PMLR}
}

@inproceedings{wu2025importance,
  title={Importance-based token merging for efficient image and video generation},
  author={Wu, Haoyu and Xu, Jingyi and Le, Hieu and Samaras, Dimitris},
  booktitle={Proceedings of the IEEE/CVF International Conference on Computer Vision},
  pages={4983--4995},
  year={2025}
}

@article{hu2024token,
  title={Token merging for training-free semantic binding in text-to-image synthesis},
  author={Hu, Taihang and Li, Linxuan and Van de Weijer, Joost and Gao, Hongcheng and Khan, Fahad S and Yang, Jian and Cheng, Ming-Ming and Wang, Kai and Wang, Yaxing},
  journal={Advances in Neural Information Processing Systems},
  volume={37},
  pages={137646--137672},
  year={2024}
}

@inproceedings{ju2023pnp,
  title={Pnp inversion: Boosting diffusion-based editing with 3 lines of code},
  author={Ju, Xuan and Zeng, Ailing and Bian, Yuxuan and Liu, Shaoteng and Xu, Qiang},
  booktitle={The Twelfth International Conference on Learning Representations},
  year={2023}
}

@article{radfordlanguage,
  title={Language Models are Unsupervised Multitask Learners},
  author={Radford, Alec and Wu, Jeffrey and Child, Rewon and Luan, David and Amodei, Dario and Sutskever, Ilya}
}

@inproceedings{kummerer2025modeling,
  title={Modeling Saliency Dataset Bias},
  author={K{\"u}mmerer, Matthias and Khanuja, Harneet Singh and Bethge, Matthias},
  booktitle={Proceedings of the IEEE/CVF International Conference on Computer Vision},
  pages={22077--22088},
  year={2025}
}

@inproceedings{parihar2025zero,
  title={Zero-Shot Depth Aware Image Editing with Diffusion Models},
  author={Parihar, Rishubh and VS, Sachidanand and Babu, R Venkatesh},
  booktitle={Proceedings of the IEEE/CVF International Conference on Computer Vision},
  pages={15748--15759},
  year={2025}
}

@inproceedings{parihar2026kontinuous,
  title={Kontinuous kontext: Continuous strength control for instruction-based image editing},
  author={Parihar, Rishubh and Patashnik, Or and Ostashev, Daniil and Radhakrishnan, Venkatesh Babu and Cohen-Or, Daniel and Wang, Kuan-Chieh Jackson},
  booktitle={Proceedings of the IEEE/CVF Conference on Computer Vision and Pattern Recognition},
  pages={37929--37939},
  year={2026}
}

@inproceedings{agrawal2026seethrough3d,
  title={Seethrough3d: Occlusion aware 3d control in text-to-image generation},
  author={Agrawal, Vaibhav and Parihar, Rishubh and Bhat, Pradhaan S and Sarvadevabhatla, Ravi Kiran and Radhakrishnan, Venkatesh Babu},
  booktitle={Proceedings of the IEEE/CVF Conference on Computer Vision and Pattern Recognition},
  pages={25403--25414},
  year={2026}
}
